\documentclass[10pt,twocolumn,letterpaper]{article}

\usepackage[pagenumbers]{cvpr} %
\usepackage[T1]{fontenc}

\definecolor{cvprblue}{rgb}{0.21,0.49,0.74}
\usepackage[pagebackref,breaklinks,colorlinks,allcolors=cvprblue]{hyperref}
\newcommand{\datasetname}{NABLA\xspace}

\def\paperID{42064} %
\def\confName{CVPR}
\def\confYear{2026}

\title{Not All Birds Look The Same: Identity-Preserving Generation For Birds}

\author{
    Aaron Sun \qquad
    Oindrila Saha \qquad
    Subhransu Maji \\
    University of Massachusetts, Amherst \\
    {\tt\small \{aaronsun, osaha, smaji\}@umass.edu}
}

\begin{document}

\twocolumn[{%
\renewcommand\twocolumn[1][]{#1}%
\maketitle\begin{center}
    \centering
    \captionsetup{type=figure}
    \vspace{-6mm}
    \includegraphics[width=0.98\linewidth]{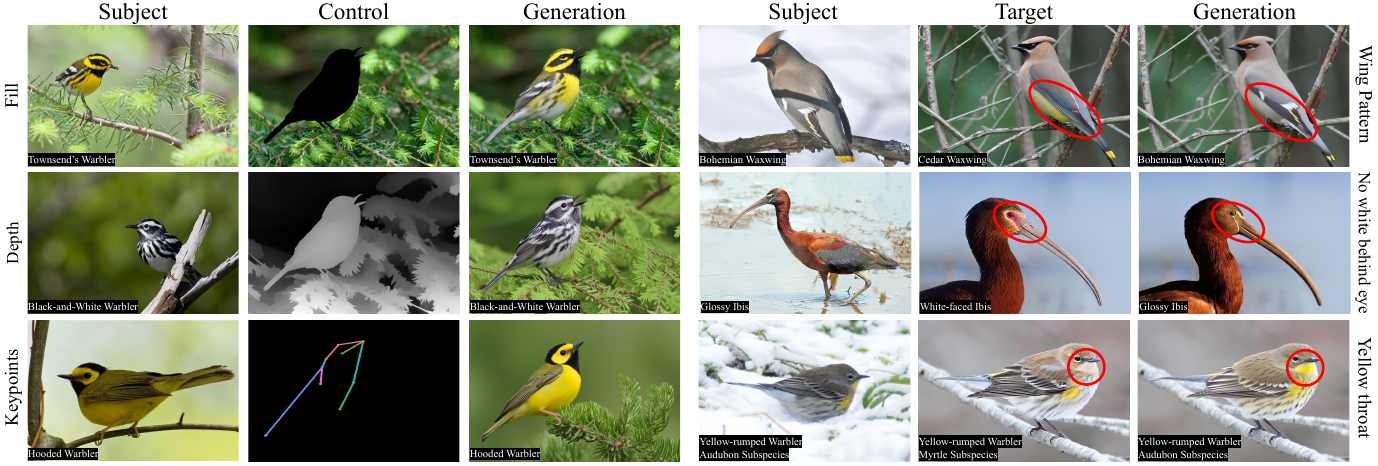}
    \vspace{-6pt}
    \captionof{figure}{
    \textbf{Qualitative results for bird generation using our method.}
    Generations match the pose of the target or control image but the appearance of the subject image.
    Left: Each row represents a different species subject being adapted into the same pose via a different control mode.
    Right: Reposing two similar species into the same canonical pose provides easier comparison between the two. 
    Our generated images exhibit greater consistency and identity preservation than existing state-of-the-art approaches (Please zoom in for details; Species names are given in the bottom left of each image).
    }
    \label{fig:fig1}
   \vspace{-2mm}
\end{center}%
}]

\begin{abstract}
Since the advent of controllable image generation, increasingly rich modes of control have enabled greater customization and accessibility for everyday users.
Zero-shot, identity-preserving models such as Insert Anything and OminiControl now support applications like virtual try-on without requiring additional fine-tuning.
While these models may be fitting for humans and rigid everyday objects, they still have limitations for non-rigid or fine-grained categories. 
These domains often lack accessible, high-quality data---especially videos or multi-view observations of the same subject---making them difficult both to evaluate and to improve upon. 
Yet, such domains are essential for moving beyond content creation toward applications that demand accuracy and fine detail.
Birds are an excellent domain for this task: they exhibit high diversity, require fine-grained cues for identification, and come in a wide variety of poses. 
We introduce the NABirds Look-Alikes~(\datasetname) dataset, consisting of 4,759 expert-curated image pairs. Together with 1,073 pairs collected from multi-image observations on iNaturalist and a small set of videos, this forms a benchmark for evaluating identity-preserving generation of birds.
We show that state-of-the-art baselines fail to maintain identity on this dataset, and we demonstrate that training on images grouped by species, age, and sex---used as a proxy for identity---substantially improves performance on both seen and unseen species.
\end{abstract}

\vspace{-18pt}
\section{Introduction}
\label{sec:intro}
Controllable image generation has advanced rapidly, offering flexible forms of high-level control---from text-to-image synthesis \cite{rombach2022high} to direct conditioning using edge maps, keypoints, or user sketches \cite{zhang2023adding}.
Recently, these models have become customizable for highly specific uses through identity-preserving generation---initially via fine-tuning on a small set of subject images \cite{ruiz2023dreambooth, gal2022image}, and more recently through zero-shot personalization from a single reference image at inference time \cite{song2025insert, tan2025ominicontrol}.
However, their usefulness beyond content creation remains limited.

\begin{figure}[t]
    \centering
    \includegraphics[width=0.45\textwidth]{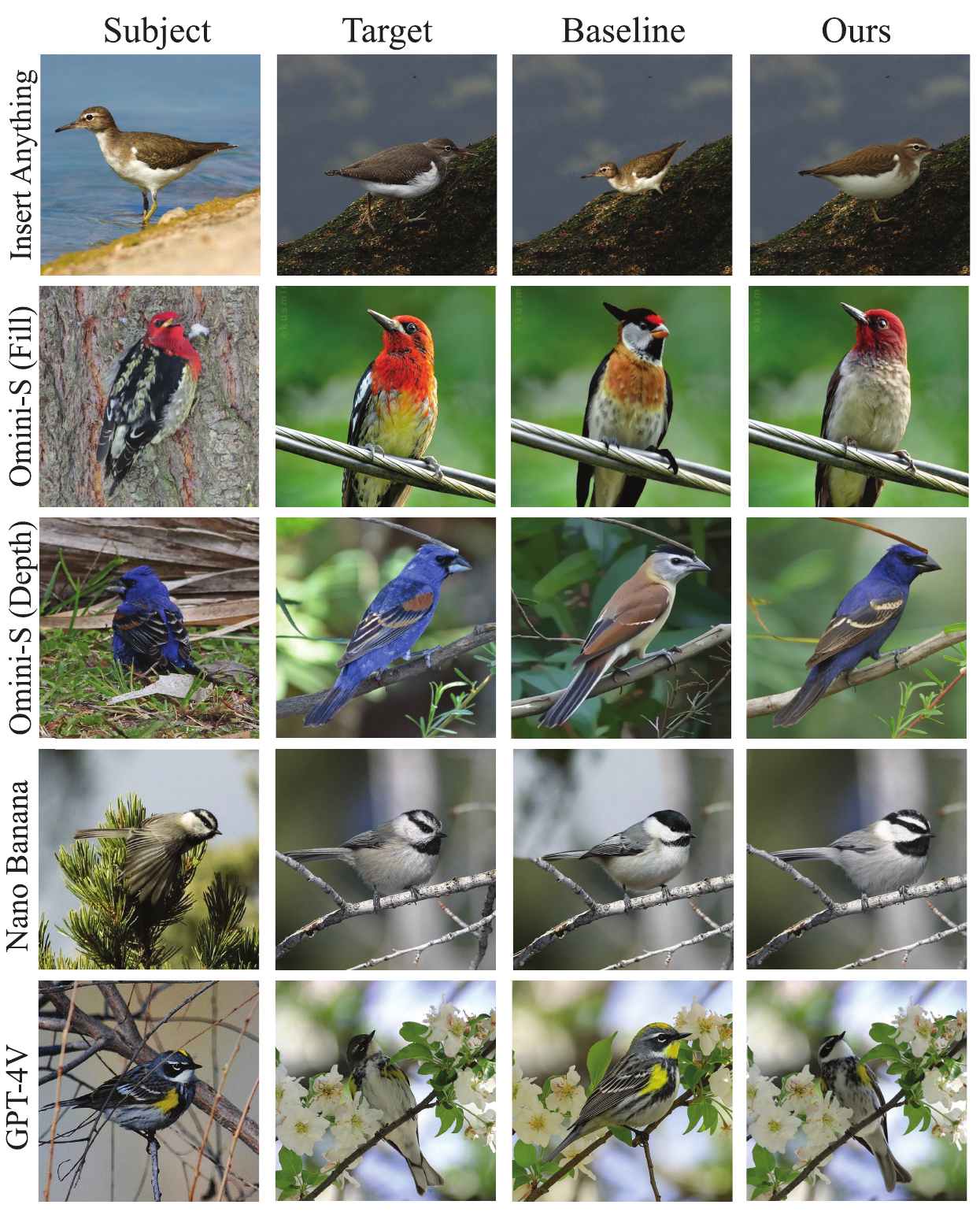}
    \vspace{-6pt}
    \caption{
        \textbf{Failure cases for baselines compared to our method.}
        Generated results should match the subject identity and target pose.
        However, the baseline models frequently change subject characteristics (rows 2, 3, 4) and target pose (rows 1, 2, 5).
        In contrast, our model generations correctly match the target image, as subject and target share an apparent identity in \datasetname.
        For the first three rows, ``Ours'' is the fine-tuned versions of the baseline models while for the last two rows we show our OminiControl with FLUX-Kontext model results. 
        Proprietary model details can be found in Appendix~\ref{sec:supp_prop_models}.
    }
    \vspace{-18pt}
    \label{fig:baseline}
\end{figure}

One particularly promising domain is fine-grained recognition, where generative models could help visualize subtle differences between instances---for example, by rendering one individual in the pose, viewpoint, or background of another to highlight discriminative features across species or even individuals, as in Figure~\ref{fig:fig1}.
Yet, current methods often fail to capture fine-grained visual details or to maintain consistency when synthesizing novel views in these domains (see Figure~\ref{fig:baseline}). %

A key challenge lies in acquiring subject-consistent training data.
Existing datasets are largely constrained to rigid objects and humans \cite{song2025insert}, or are entirely AI-generated \cite{tan2025ominicontrol}.
Comparable datasets for natural fine-grained categories are scarce, limiting progress in applications where accuracy and attention to detail are critical---such as scientific and educational visualization.

We address this gap by introducing a benchmark for identity-preserving image generation in the fine-grained domain of birds.
Birds pose distinctive challenges absent from existing benchmarks due to their non-rigid structures and elaborate, class-specific visual appearance.
To mitigate the lack of large-scale, subject-consistent data, we leverage the taxonomic structure of fine-grained categories to construct bird look-alikes---pairs sharing the same species, age, sex, and seasonal variant that exhibit strong visual similarity.
For evaluation, we curate additional pairs from the NABirds dataset~\cite{van2015building} through expert annotation and compile a test set from multi-image observations in iNaturalist~\cite{iNaturalist}, where individual identity is known to be preserved.
We show that training within this framework improves image generation quality across both seen and unseen bird species, demonstrated qualitatively (Figures~\ref{fig:fig1}, \ref{fig:success}, \ref{fig:inat_unseen}) and quantitatively using standard identity-preserving metrics (Table~\ref{tab:results}).
Our framework, built on OminiControl~\cite{tan2025ominicontrol} and Insert Anything~\cite{song2025insert}, enables control through masks, depth maps, and keypoints, in addition to text.
The dataset and benchmarking code are available at \url{https://github.com/cvl-umass/nabla}. 

In summary, our main contributions are as follows:
\begin{enumerate}
\item We propose NABirds Look-Alikes (\datasetname), an evaluation benchmark of 4759 hand-selected image pairs across 401 species for the task of identity-preserving image generation of birds.
\item We show that metrics on \datasetname correlate strongly with true identity pairs extracted from multi-image iNaturalist observations, and that images in \datasetname are of consistently higher quality than existing alternatives. 
\item We demonstrate that training on image pairs using species, age, and sex as proxies for identity leads to improved generation fidelity and quality, achieving a 41\% reduction in MSE on \datasetname over the baseline model and showing strong generalization to unseen species.
\item Finally, we showcase applications of this framework in visualization and machine teaching. %
\end{enumerate}

\section{Related Work}
\label{sec:related}

\subsection{Identity-Preserving Generation}
Recent improvements on controlling diffusion models have focused on identity-preservation, meaning generations maintain the identity of a specific object or individual. 
DreamBooth~\cite{ruiz2023dreambooth} and Textual Inversion~\cite{gal2022image} proposed frameworks using a small set of subject images to fine-tune the diffusion model or the subject token, respectively. 
In contrast, a recent line of works has focused on zero-shot identity-preserving generation based on a single subject image.
AnyDoor \cite{chen2024anydoor} fine-tunes an identity and detail extractor to generate features that are injected into a diffusion model.
Insert Anything \cite{song2025insert} specifies the task using multi-panel images called ``polyptyches'' with subject and background panels, but is restricted to inpainting as the control mode. 
OminiControl \cite{tan2025ominicontrol} and DreamO \cite{mou2025dreamo} allow for multiple conditioning types to be mixed and matched by directly inputting the conditioning images as tokens to the diffusion model.
These zero-shot models all train on a variety of datasets derived from videos, multi-view images, and synthetic generations, focusing on virtual try-on, everyday objects, and style transfer with existing benchmarks reflecting these tasks \cite{peng2024dreambench, ge2019deepfashion2, sundarampersonalized, kotar2023these}.
Closed-source models such as GPT-4V~\cite{gpt4v} and Nano Banana~\cite{nanobanana} likely built on these frameworks, but their training methods and datasets are unknown.
This approach may be effective for creative applications, but suffers for scientific applications where data is scarce and accuracy is paramount (see Figure~\ref{fig:baseline}).
\vspace{-3pt}
\subsection{Multi-view Bird Datasets}
\begin{figure}[ht]
    \centering
    \includegraphics[width=0.46\textwidth]{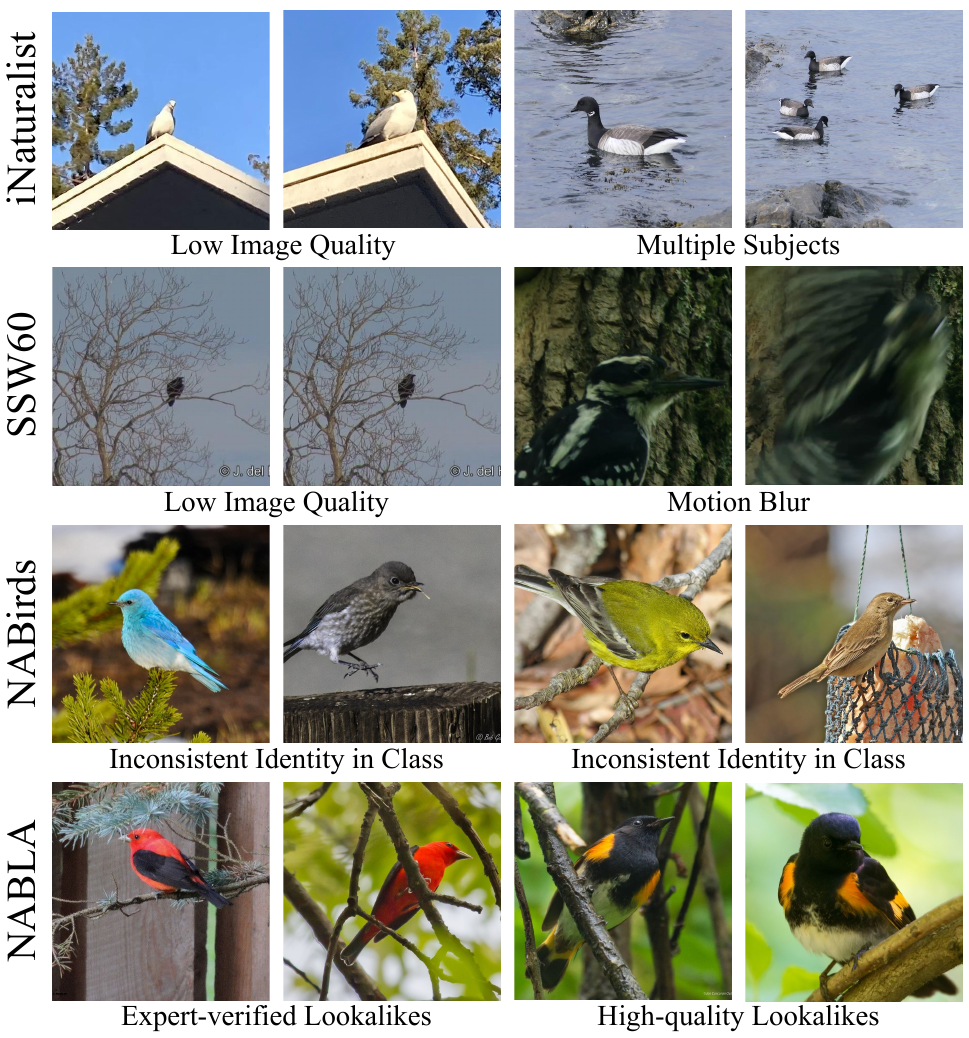}
    \vspace{-6pt}
    \caption{
    \textbf{Image pair examples from 4 datasets.}
    Though iNaturalist and SSW60 have true identity-preservation, other qualities such as image quality and motion blur make them poor for image generation.
    NABirds consists of single-subject, high-quality images, but has inconsistent identity, even within classes.
    In contrast, \datasetname has expert-verified lookalike bird pairs on high-quality NABirds images.
    }
    \vspace{-15pt}
    \label{fig:ssw_inat}
\end{figure}

Although many existing datasets have multi-view information for individual birds, these datasets often have lower image quality and species diversity when compared to standard image classification datasets like CUB200~\cite{WahCUB_200_2011} and NABirds~\cite{van2015building}.
The difficulty of collecting true multi-view data of a single subject limits these environments to cages or aviaries with usually a single species \cite{badger20203d, naik20233d, moradi2025context}.
In contrast, single-view video datasets have significantly higher species and environment diversity \cite{ssw602022eccv, rodriguez2025visual, ge2016temporal, 7552915, sun2024fbdsv2024, Sun_2022_ACCV, hagerlind2024temporally}, but are typically lower quality than their image counterparts due to motion blur and difficulty in data collection (see Figure~\ref{fig:ssw_inat}).
Meanwhile, existing 3D animal datasets focus on 4-legged mammals \cite{xu2023animal3d, Zuffi:CVPR:2017}, whose deformations have a less wide range of locomotive needs compared to birds (e.g., perching, swimming, and flying), or are synthetically generated \cite{li20214dcomplete, objaverseXL}.
While WildlifeReID~\cite{adam2025wildlifereid} and PetFace~\cite{shinoda2024petface} both provide multi-image views of the same subject, these datasets are limited in terms of image resolution, pose and lighting variation, and species diversity, especially for image generation.
Separately from formal datasets, citizen scientists on iNaturalist~\cite{iNaturalist} and eBird~\cite{sullivan2009ebird} provide multi-image collections of the same individual but these can vary in quality (Figure~\ref{fig:ssw_inat}).
Broadly speaking, no dataset with same-subject bird images exists at the scale and quality required for generative training.
However, just as previous works on fine-grained domains have overcome these hurdles by using coarse data~\cite{saha2024improved, saha2025generate}, we find training using matching species, age, and sex pairs improves performance on true same-identity pairs.
Additional datasets are discussed in Appendix~\ref{sec:supp_additional_datasets}.

\subsection{Fine-grained Generation}
Despite the lack of explicit evaluation data, 3D mesh reconstructions for birds based on images has been explored in considerable depth. 
These works typically train from image collections of given species as well as other helpful information such as subject mask, keypoints, part segmentation, and even articulated skeletons \cite{kanazawa2018learning, li2020self, goel2020shape, wang2021birds, wu2023magicpony}.
To bypass the lack of ground truth 3D shapes, these works typically evaluate on proxy tasks such as camera pose, mask, or keypoint prediction.
These 3D reconstructions have been utilized for classification \cite{joung2021learning}, style transfer \cite{wang2023creative}, and tracking \cite{hagerlind2024temporally}.
Existing works for diffusion on fine-grained classes focus on improving class-conditioned generation \cite{khurana2024hierarchical, monsefi2025taxadiffusion, pan2025finediffusion, ng2024chirpy3d}, but this lacks identity-preservation or fine pose controls. 
While class-conditioned generation offers hierarchical structure and taxonomic information, this conditioning does not extend well to unseen species and abnormal or drab individuals.
DIFFusion~\cite{chiquier2025teaching} showed how inter-species generation which we emulate in Figure~\ref{fig:fig1} can be used for machine teaching, but this method lacks pose controls and identity-preservation.

\section{Methods}
\label{sec:methods}
\begin{figure*}[ht]
    \centering
    \includegraphics[width=0.92\textwidth]{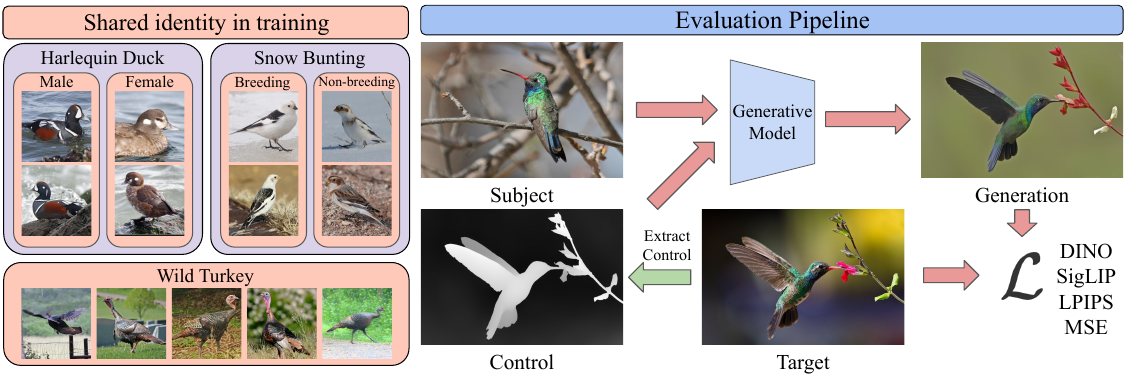}
    \vspace{-3pt}
    \caption{
   \textbf{ Dataset usage and pipeline.}
    Left: In training, only images within the same class (shown in pink) are considered as pairs for sampling. 
    Classes can vary in their hierarchy (species-level, gender-level, etc.) dependent on species.
    Right: During evaluation, the subject image and the control of the target image are inputted to the model for generation. 
    The generation and target image are masked and the birds are evaluated using DINO, SigLIP, LPIPS, and MSE.
    }
    \vspace{-15pt}
    \label{fig:pipeline}
\end{figure*}

\subsection{Evaluation Datasets}

\noindent\textbf{NABirds Look-Alikes (\datasetname).}
We asked a small group of bird experts to manually annotate the NABirds test set with the following task: given an image of a bird, select another image from the same class where the individual appears the same as in the other image. 
The annotators chose anywhere from 5-10 image pairs per class, if possible, where no images are repeated. 
Some classes had too much individual variation to select 10 pairs.
Through this method, we created the NABirds Look-Alikes (\datasetname) dataset, a test set of 4759 image pairs which share an apparent identity while maintaining a standard of single-subject, high-quality images suitable for generation (Figure~\ref{fig:ssw_inat}).
Exact annotation protocol, dataset statistics, and more examples are given in Appendix~\ref{sec:supp_datasets}.
Though not perfect, \datasetname mimics true shared-identity data which does not currently exist at the scale and quality necessary for training and evaluation.

\noindent\textbf{iNaturalist Image Sets.}
We also acquired ground truth test pairs from iNaturalist~\cite{iNaturalist}. 
On iNaturalist, citizen scientists upload photos of species observations, where each observation represents a single individual.
As a result, identity is preserved between images within the same observation. 
However, in practice we found these images can be low quality or contain multiple birds (Figure~\ref{fig:ssw_inat}).
We downloaded 677 image pairs from species in the NABirds dataset and 396 image pairs from species outside the dataset, creating iNat-Seen and iNat-Unseen, respectively.
Additional information and examples are given in Appendix~\ref{sec:supp_datasets}.

\noindent Lastly, we present additional results on WildlifeReID-10k~\cite{adam2025wildlifereid}, an animal re-identification dataset, in Section~\ref{sec:supp_indbirds}.

\subsection{Evaluation Metrics}
For each test pair, metrics are calculated with the first image as the subject and the second image as the target and vice versa.
After generation, both the generated image and target image are masked using the subject mask to remove background contributions. 
This evaluation pipeline is also shown in Figure~\ref{fig:pipeline}. 
We evaluate on 4 metrics: DINOv2 \cite{oquab2023dinov2} feature similarity, SigLIP feature similarity \cite{zhai2023sigmoid}, Learned Perceptual Image Patch Similarity (LPIPS) \cite{zhang2018perceptual}, and mean squared error (MSE).
MSE and LPIPS are effective for comparing overall image similarity. 
Unlike most other generation tasks, we also use DINOv2 and SigLIP feature similarity to evaluate how class-level features and pose information are retained between the target and generation.

\subsection{Control Methods}
\begin{figure}[t!]
    \centering
    \includegraphics[width=0.47\textwidth]{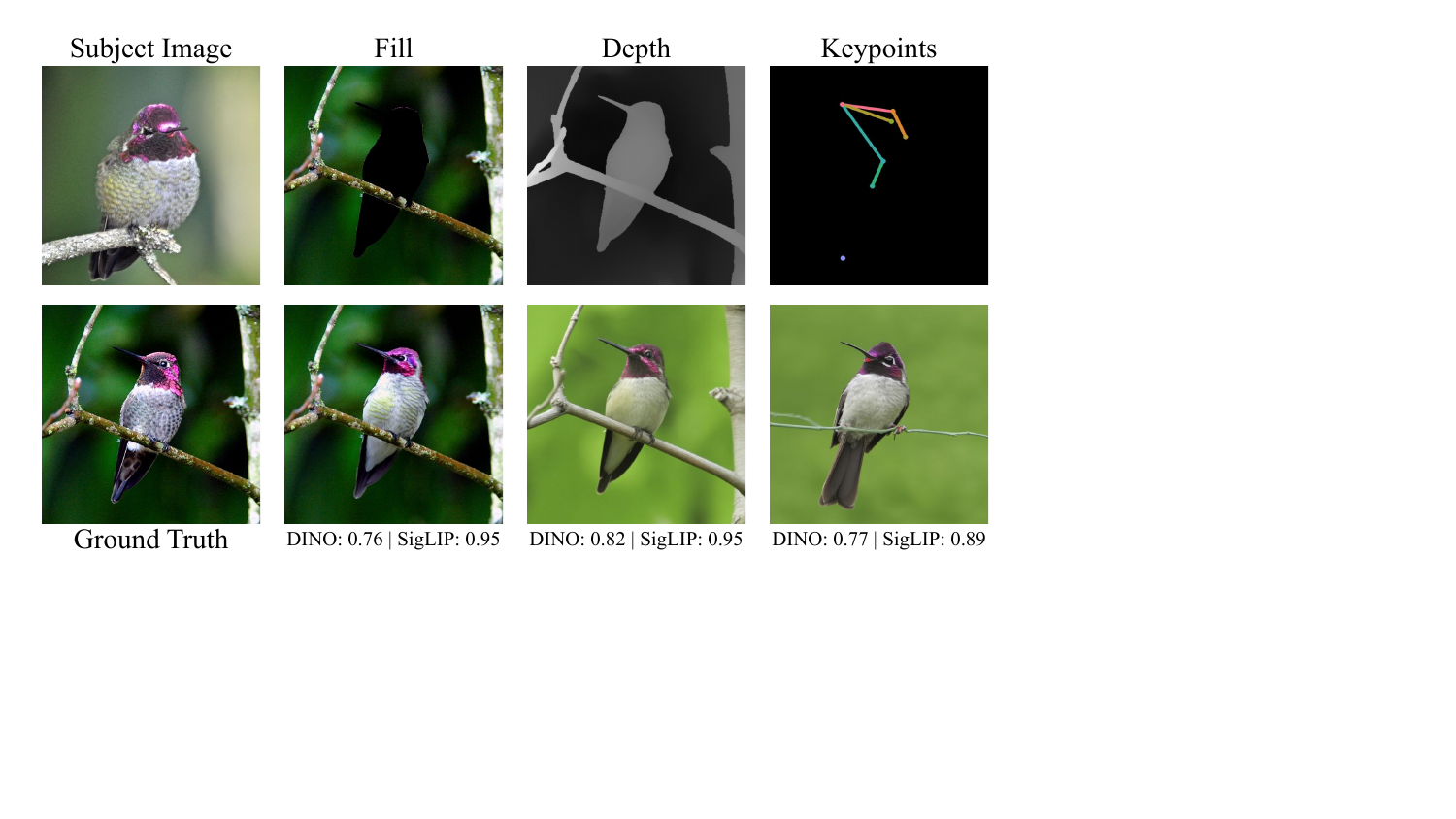}
    \vspace{-3pt}
    \caption{
    \textbf{Varying control mode settings for training and evaluation.} The top row shows model inputs and the bottom row shows fine-tuned model outputs and scores.
    }
    \vspace{-15pt}
    \label{fig:control}
\end{figure}

\noindent\textbf{Fill.} 
Fill represents the generation task as an inpainting task.
The control image is given as the target image with the subject removed, as in Figure~\ref{fig:control}. 
These masks are acquired using SAM2 \cite{ravi2024sam2} with bounding box annotations for NABirds and Grounded-SAM-2 \cite{ren2024grounded} with the text prompt ``bird'' for iNaturalist images. 
For models which require a text prompt, we use ``an image of a bird'' for training and evaluation.

\noindent\textbf{Depth.} 
The depth control mode provides the depth map of the target image. 
These depth maps are generated using Video-Depth-Anything \cite{video_depth_anything} in image mode.
Since this leads to ambiguity in the background, we provide a basic caption of the background discussed below and in Appendix~\ref{sec:supp_caption}.
Depth gives slightly stronger control over bird pose as opposed to mask-based control which can lead to pose ambiguities, as shown in Figure~\ref{fig:failure}. 

\noindent\textbf{Keypoints.}
We lastly define a generation task in terms of keypoints. 
NABirds provides 11 keypoint annotations per image for various body parts including bill, left eye, belly, etc. 
We generate a skeleton control image using these keypoints with uniquely colored joints and connections, as in Figure~\ref{fig:control}.
We also provide a caption as in the depth task.
Keypoints are not readily available for iNaturalist data, so we exclude this mode of control for evaluation on those datasets. 
Further information is given in Appendix~\ref{sec:supp_keypoint}.

\noindent\textbf{Captions.} 
Since OminiControl~\cite{tan2025ominicontrol} requires text prompts for generation and the depth and keypoint tasks introduce ambiguity in the background, we add captions to these tasks to give background information. 
We generate these captions in two modes, short and long captions, using Qwen2.5 VL \cite{Qwen2.5-VL}.
The details of the prompts used and example captions are in Section~\ref{sec:supp_caption}.

\subsection{Proxy Identity Training}
We fine-tune on the training set of NABirds for this task. 
At each training step, two images of the same class (species, age, sex, etc.) are sampled at random. 
These classes can vary in specificity as shown in Figure~\ref{fig:pipeline}. 
This sampling method generates coarse pairs for training but can still produce mismatches such as in Figures~\ref{fig:ssw_inat} and \ref{fig:supp_mismatch}, but we expect this to be a reasonable proxy on average.
One of the two images is designated at random as the subject image and the other is the target image.
The control image (mask and background, depth, or keypoints) and caption (if necessary) are extracted from the target image, then the subject and control images are used for training the generation model. 

We follow the training processes outlined in the Insert Anything \cite{song2025insert} and OminiControl \cite{tan2025ominicontrol} architectures.
For Insert Anything, subject and masked background images were stitched into a diptych panel and a FLUX.1-Fill backbone is fine-tuned with LoRA. 
We followed the mask augmentation procedure used for objects, randomly augmenting with Bessel curve-based shapes and bounding boxes. 
For OminiControl, the subject and condition latent tokens are appended to the noisy latent tokens before going through a DiT \cite{peebles2023scalable}.
The control image utilized a spatially-aware encoding while the subject image did not.
Each control mode was trained separately and on two backbones, FLUX.1-Schnell (Omini-S) and FLUX.1-Kontext (Omini-K) \cite{labs2025flux}.
To ensure high quality results, we use input/output images of size 1024x1024 and train on 4 A100/H100 GPUs for 10000 iterations, which takes about 3 days.
For more details, see Appendix~\ref{sec:supp_hyperparameters}.

\begin{figure}[h!]
    \centering
    \includegraphics[width=0.47\textwidth]{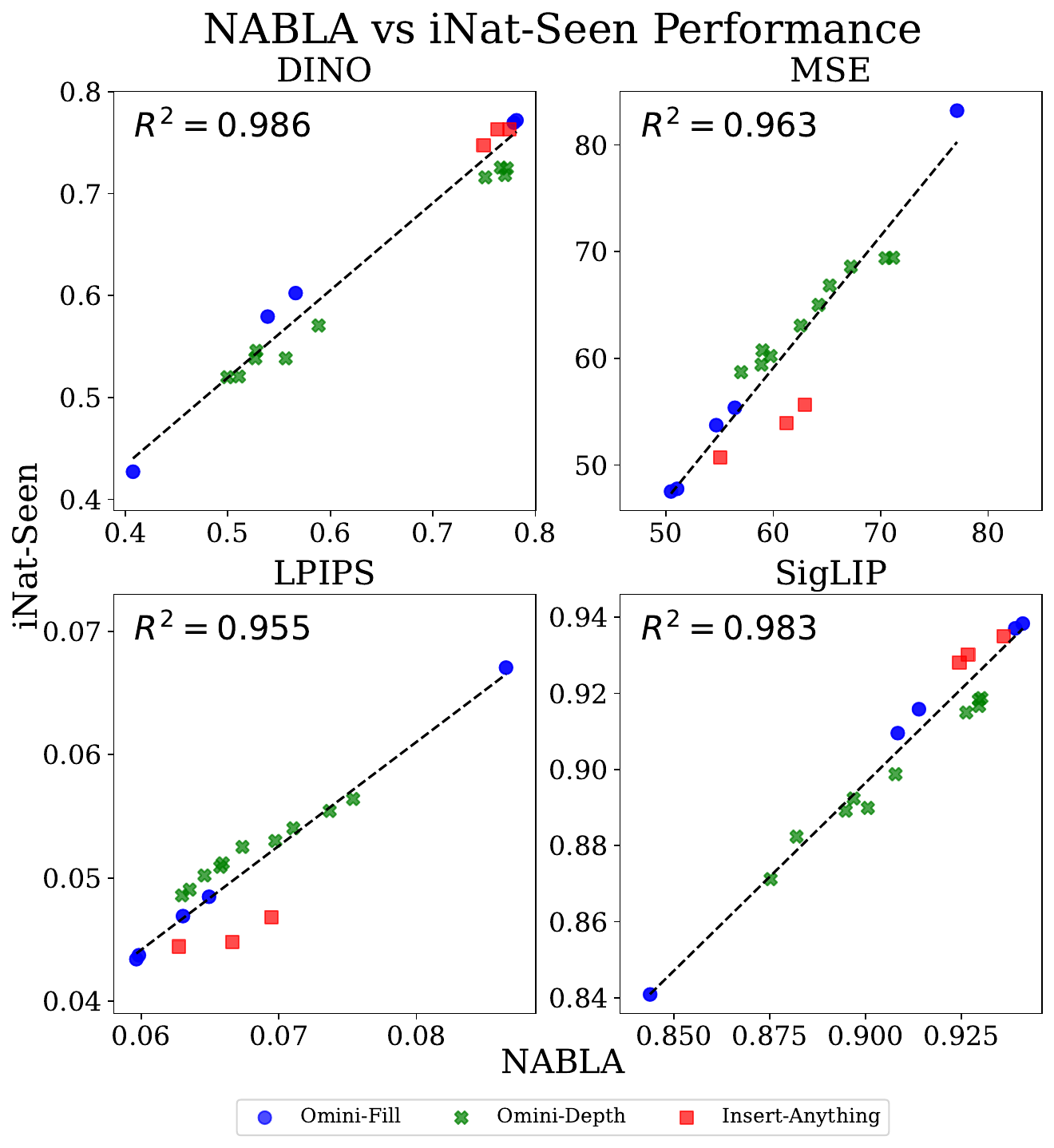}
    \caption{
    \textbf{Comparison of average model performance on \datasetname and iNat-Seen data across all trials.}
    Performance on \datasetname is very indicative of performance on iNat-Seen.
    Similar graphs are given for the other dataset pairs in Appendix~\ref{sec:supp_correlation_graphs}.
    }
    \vspace{-9pt}
    \label{fig:graph_nabla_inat}
\end{figure}

\begin{table*}[h!]
  \centering
  \resizebox{0.93\textwidth}{!}{
  \begin{tabular}{l l | c c c c | c c c c | c c c c}
  \toprule
  \multicolumn{2}{c|}{} & \multicolumn{4}{c|}{\datasetname} & \multicolumn{4}{c|}{iNat-Seen} & \multicolumn{4}{c}{iNat-Unseen} \\
  Control & Arch & DINO $\uparrow$ & SigLIP $\uparrow$ & LPIPS $\downarrow$ & MSE $\downarrow$ & DINO $\uparrow$ & SigLIP $\uparrow$ & LPIPS $\downarrow$ & MSE $\downarrow$ & DINO $\uparrow$ & SigLIP $\uparrow$ & LPIPS $\downarrow$ & MSE $\downarrow$ \\
  \midrule

  \multirow{3}{*}{Depth} 
  & Om-S* & 0.50 & 0.88 & 0.074 & 70.4 & 0.52 & 0.88 & 0.055 & 69.4 & 0.55 & 0.88 & \underline{0.038} & 68.1 \\
  & Om-S & \underline{0.59} & \underline{0.91} & \underline{0.066} & \underline{62.5} & \underline{0.57} & \underline{0.90} & \underline{0.051} & \underline{63.1} & \underline{0.61} & \underline{0.90} & \textbf{0.034} & \underline{62.0} \\
  & Om-K & \textbf{0.77} & \textbf{0.93} & \textbf{0.063} & \textbf{57.0} & \textbf{0.72} & \textbf{0.92} & \textbf{0.049} & \textbf{58.7} & \textbf{0.72} & \textbf{0.91} & \textbf{0.034} & \textbf{59.4} \\
  \midrule
  
  \multirow{5}{*}{Fill}
  & Om-S* & 0.41 & 0.84 & 0.087 & 77.1 & 0.43 & 0.84 & 0.067 & 83.2 & 0.49 & 0.85 & 0.046 & 80.8 \\
  & Om-S & 0.57 & 0.91 & \underline{0.063} & \underline{54.7} & 0.60 & 0.92 & 0.047 & 53.7 & 0.65 & 0.92 & 0.031 & 52.6 \\
  & Om-K & \textbf{0.78} & \textbf{0.94} & \textbf{0.060} & \textbf{51.0} & \textbf{0.77} & \textbf{0.94} & \textbf{0.043} & \textbf{47.8} & \textbf{0.78} & \textbf{0.94} & \textbf{0.029} & \textbf{46.9} \\
  & Ins-A* & 0.75 & \underline{0.92} & 0.069 & 62.9 & 0.75 & \underline{0.93} & 0.047 & 55.6 & 0.76 & \underline{0.93} & 0.031 & 54.4 \\
  & Ins-A & \underline{0.77} & \textbf{0.94} & \underline{0.063} & 55.0 & \underline{0.76} & \textbf{0.94} & \underline{0.044} & \underline{50.7} & \underline{0.77} & \underline{0.93} & \underline{0.030} & \underline{51.1} \\
  \midrule
  
  \multirow{2}{*}{Keypoint}
  & Om-S & \underline{0.65} & \textbf{0.91} & \textbf{0.075} & \underline{68.0} & {---} & {---} & {---} & {---} & {---} & {---} & {---} & {---} \\
  & Om-K & \textbf{0.69} & \textbf{0.91} & \underline{0.076} & \textbf{66.7} & {---} & {---} & {---} & {---} & {---} & {---} & {---} & {---} \\
  
  \bottomrule
  \end{tabular}}%
  \caption{
  \textbf{Comparison of model performance across datasets and control modes.} * indicates baseline, Ins-A stands for InsertAnything, and Om-S and Om-K stand for OminiControl with FLUX-Schnell and FLUX-Kontext backbones, respectively. 
  The best result in each control/metric is \textbf{bolded} and the second highest is \underline{underlined}.
  The baseline model for Om-K* does not exist and keypoints are not available for iNat data, so those entries are excluded.
  Fine-tuning improves performance consistently across all three datasets.
  }
  \label{tab:results}
\end{table*}

\begin{figure*}[h!]
    \centering
    \includegraphics[width=0.94\textwidth]{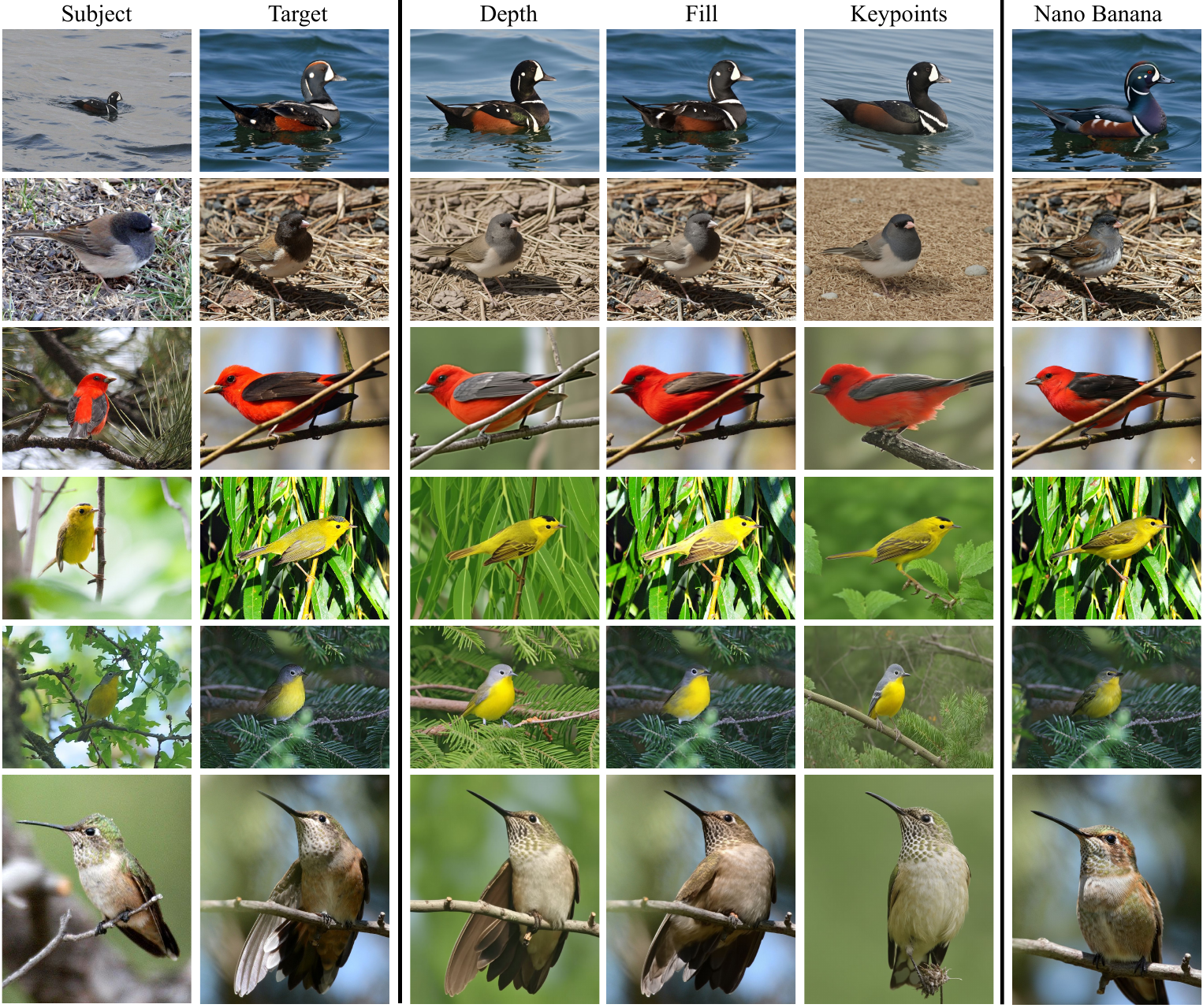}
    \caption{
\textbf{    Generation results on \datasetname for our best fine-tuned models in each control method vs inpainting Nano Banana.}
    Our models are generally more faithful to subject characteristics and target pose.
    We highlight discrepancies in Nano Banana generations: 
    Row 1 shows 2 cheek spots instead of 1.
    Row 2 shows an altered ``bib'' pattern.
    Row 3 and 4 have the wrong head and wing position, respectively.
    Row 5 shows an eyebrow which the subject does not have.
    Row 6 has the entirely wrong pose.
    }
    \label{fig:success}
\end{figure*}

\begin{figure*}[h!]
    \centering
    \includegraphics[width=0.99\textwidth]{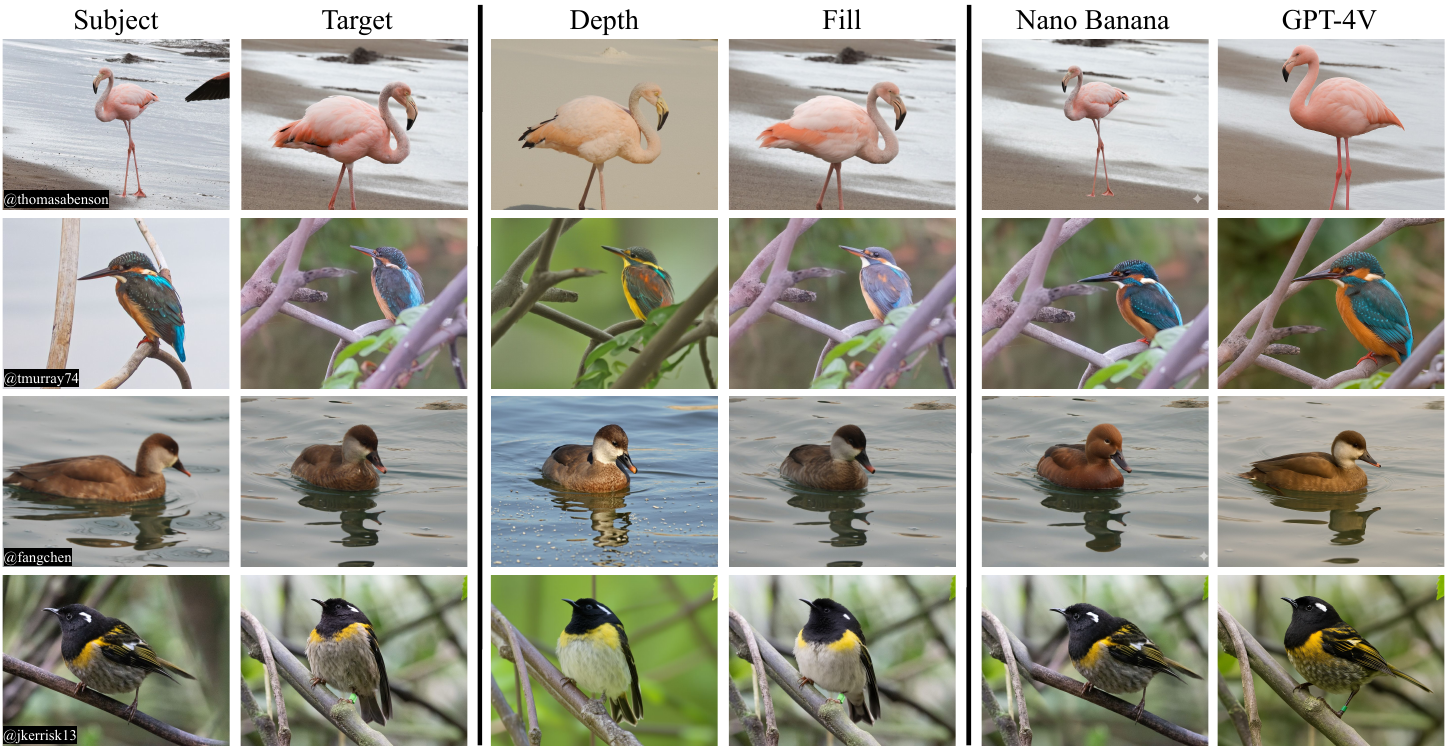}
    \vspace{-6pt}
    \caption{
    \textbf{Generation results on iNat-Unseen for our best fine-tuned models in each control method along with proprietary baselines.}
    Despite being trained on NABirds, our models appear to generalize to unseen species. 
    The proprietary models are inconsistent in reflecting subject identity and target pose. 
    Images are from iNaturalist~\cite{iNaturalist} with Observer IDs listed.
    }
    \vspace{-14pt}
    \label{fig:inat_unseen}
\end{figure*}

\section{Results}
\label{sec:results}

\subsection{Quantitative Results}
\label{sec:quantitative}
We begin by examining quantitative results across \datasetname and iNaturalist data in Table~\ref{tab:results}. 

\noindent\textbf{Performance on \datasetname correlates strongly with performance on ground truth identity pairs.}
As seen in Figure~\ref{fig:graph_nabla_inat}, across all metrics and settings we see very high correlation between performance on \datasetname and true matching identity pairs on iNaturalist. 
Clearly, \datasetname is effective in measuring identity-preservation while not having undesirable qualities associated with ground truth identity pairs (Figure~\ref{fig:ssw_inat}).
While DINO, MSE, and SigLIP show a near one-to-one correlation, LPIPS is slightly lower in \datasetname compared to iNat-Seen.
We attribute this to lower quality and smaller subjects in iNaturalist data compared to NABirds, meaning details are less apparent and less impactful on performance.
We also saw similar improvement trends on the Individual-Birds~\cite{ferreira2020deep} in WildlifeReID-10k~\cite{adam2025wildlifereid} which we present in Appendix~\ref{sec:supp_indbirds}.

\noindent\textbf{Training on NABirds improves performance across all test sets, including on unseen species.}
Though our training data uses species, age, sex, and breeding status as a proxy for identity, we see this is still effective for improving identity-preservation on this task.
As seen in Table~\ref{tab:results} and Figure~\ref{fig:baseline}, we observe performance improvements compared to baseline models from OminiControl \cite{tan2025ominicontrol} and Insert Anything \cite{song2025insert} on all three datasets and control modes.
Though Insert Anything is a fairly strong baseline, we find our fine-tuned Insert Anything and Omini-K both outperform this baseline across the board. 
We also find Omini-K to be slightly better than Insert Anything for inpainting while retaining versatility from its mixed control capabilities. 
Impressively, training on NABirds also generalizes to unseen species.
Improvements on iNat-Seen and iNat-Unseen are similar, indicating the model improves on identity-preserving bird generation as a whole rather than just for species in NABirds.
Although this training is effective, we highlight that proxy identity is insufficient for evaluation where accuracy and attention to detail are paramount as these proxies have mismatches (Figures~\ref{fig:ssw_inat} and \ref{fig:supp_mismatch}).

\noindent\textbf{Inpainting performs slightly better than depth map control on our task.}
We found this result to be unintuitive, since masks are insufficient for specifying pose and leads to ambiguities seen in Figure~\ref{fig:failure}. 
Despite that, we see the best-performing inpainting models outperform their depth-based counterparts on all datasets.
We attribute this to two possible factors. 
Firstly, in training we do not modify the diffusion loss, meaning the unspecified background may impact the training loss which may have undesirable effects on training.
In addition, during evaluation we see in Figure~\ref{fig:success} that lack of background context can lead to different lighting on the subject itself, which in turn can influence performance.
These problems are only amplified for keypoint-based control, leading to worse performance than inpainting and depth-based controls. 

\subsection{Qualitative Results}
\subsubsection{Intra-species Results}
\noindent\textbf{Trained models retain identity and match pose much better than the baselines.}
As shown in Figure~\ref{fig:baseline}, we see baseline models frequently fail to accurately follow task restrictions.
While Insert Anything reasonably retains class properties, it can fail to reflect the target pose which is rectified after training (Figure~\ref{fig:baseline}).
Similarly, closed-source models such as Nano Banana \cite{nanobanana} and GPT-4V \cite{gpt4v} have pitfalls with respect to reflecting target pose and maintaining species-level features. 
In contrast, we see in Figures~\ref{fig:baseline}, \ref{fig:success}, and \ref{fig:inat_unseen} that trained models reflect both subject identity and pose constraints more faithfully than baseline models.

\begin{figure}[h!]
    \centering
    \includegraphics[width=0.47\textwidth]{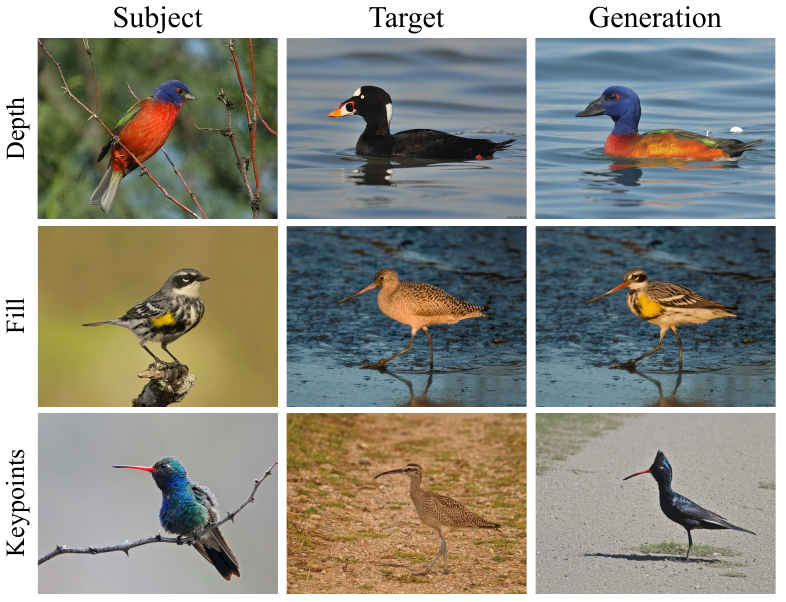}
    \caption{
    \textbf{Inter-species results.}
    Inter-species generations where the subject and target have significantly different shapes can lead to unrealistic, but creative results.
    }
    \vspace{-18pt}
    \label{fig:mixed_species_fail}
\end{figure}

\subsubsection{Inter-species Results}
\noindent\textbf{Generation capabilities allow for inter-species control.}
We observe in Figure~\ref{fig:fig1} we are able to generate results in a cross-species fashion for similarly-shaped species. 
Despite lack of ground-truth pairs, we see reasonable generations across a variety of species.
While the requirement of similar species shape can be restrictive for unique animals, the diversity of bird species gives a vast array of species within the same family and genus for inter-species control.
We could envision this being useful for endangered or rare species: a limited set of images could be used in tandem with control images of a related species to generate numerous images in a realistic fashion. 
Regardless, the requirement for ground-truth images of similar species can still be restrictive, which we discuss further in Section~\ref{sec:model_pitfalls}. 
We find in Figure~\ref{fig:mixed_species_fail} that using distant species as control images leads to creative, but unrealistic results.

\noindent\textbf{Cross-species control enables easy comparisons.}
An obvious benefit of cross-species control capabilities is much clearer direct comparison between species or even individuals of the same species. 
We see in Figure~\ref{fig:fig1} that distinguishing characteristics become more apparent in a ``side-by-side'' view compared to when the two individuals have significantly different poses. 
We anticipate this could be helpful for scientific or educational purposes, such as for individual re-identification or for highlighting differences between species. 
This method of machine teaching was quantitatively shown to be effective in DIFFusion~\cite{chiquier2025teaching}.

\vspace{-3pt}
\subsection{Limitations}
\label{sec:model_pitfalls}
\begin{figure}[ht!]
    \centering
    \includegraphics[width=0.44\textwidth]{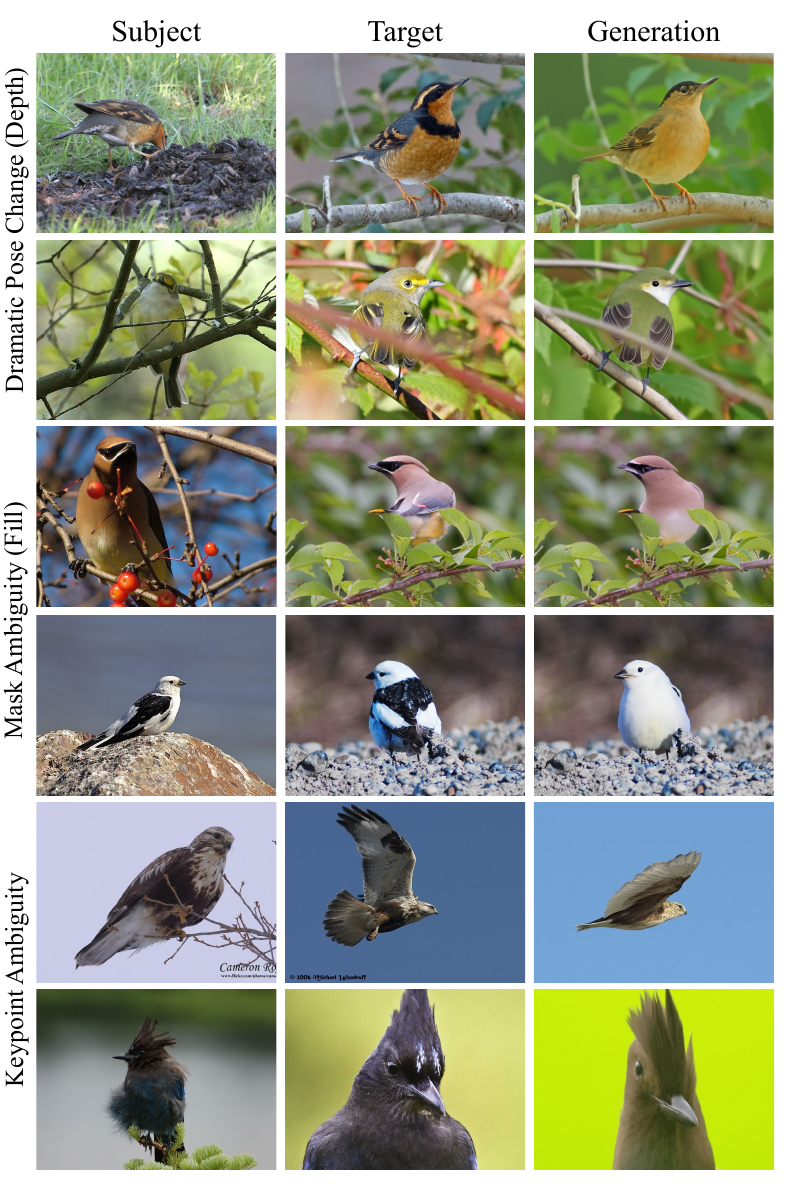}
    \caption{
    \textbf{Failure cases for fine-tuned models.}
    Depth models still fail on drastic pose changes (rows 1 and 2) and inpainting and keypoint suffers from control ambiguity (rows 3 through 6). 
    More explicitly, row 1 lacks the black collar and eyemask, row 2 lacks the white eye, rows 3 and 4 are facing the wrong way, and rows 5 and 6 are misproportioned.
    }
    \vspace{-18pt}
    \label{fig:failure}
\end{figure}

\vspace{-3pt}
\noindent\textbf{Fill models and depth models both have weaknesses (Figure~\ref{fig:failure}).}
Fill models and depth models lack specificity to accomplish the identity-preserving generation task for birds completely.
Fill models suffer from pose ambiguity from a given mask, due to silhouettes being insufficient to define pose. 
In contrast, while depth-based models have additional pose information, they suffer from undesired background effects in training and evaluation discussed in Section~\ref{sec:quantitative} and are still unable to handle large pose changes.
We see room for improvement in this area with respect to problem specification. 
Clearly, combining depth and fill control could mitigate each mode's weaknesses.
Yet, we found a simple combination of depth + fill led to little improvement (Appendix~\ref{sec:supp_mixed}).
This form of control is also hyper-restrictive, making it essentially infeasible for use without ground-truth reference images.
We envision the best model would be easy to control without ground-truth images while still being capable of realistic generation when given reference images.

\noindent\textbf{Keypoint-based control is far inferior to inpainting and depth map controls.}
Since fill and depth models are very restrictive in their usage, we tried training a keypoint-based model which could be manipulated easily for creative uses.
Unsurprisingly, this type of model was much less effective than other modes of control after training.
Keypoint-based models suffer even more strongly from the drawbacks associated with depth models which led to more ineffective training.
The solution for easily manipulatable control remains unclear for birds, whose deformable properties make them harder to specify pose than humans which conform more easily to a skeleton.  

\noindent\textbf{Species properties are not retained for difficult examples.}
We see in Figure~\ref{fig:failure} that implicit class information is not effectively learned by the model with our basic training. 
Theoretically, a generative model which classifies the subject species then generates an image based on this classification would be able to handle drastic pose changes.
Yet, the method for effectively enforcing this during training while extending to unseen species is unclear, and we see it as a clear future direction of improvement.

\section{Conclusion}
\label{sec:conclusion}
\vspace{-3pt}
Despite recent improvements in image generation, the use of these models remains limited mostly to content creation.
This originates from an inability to verify the fidelity of image generations, particularly for scientific applications where accuracy is crucial. 
We propose a method to evaluate and improve identity-preserving generation on birds, a particularly difficult group due to their fine-grained details and vast range of deformations.
To combat lack of appropriate same-subject data we create \datasetname, a high-quality look-alikes benchmark which mirrors performance on true same-subject pairs.
We improve baseline model performance on \datasetname by fine-tuning on NABirds using species, age, and sex as a proxy for identity. 
However, despite this progress, this task remains unsolved in many respects.
We hope future work will explore machine teaching in the context of bird identification, address the weaknesses of existing control modes, and improve performance on difficult examples.

\section*{Acknowledgments}
This project was supported in part by National Science Foundation Grant \#2329927 and a NAIRR compute allocation. We also thank the users of iNaturalist for sharing their observations online.

{
    \small
    \bibliographystyle{ieeenat_fullname}
    \bibliography{main}

\begin{thebibliography}{59}
\providecommand{\natexlab}[1]{#1}
\providecommand{\url}[1]{\texttt{#1}}
\expandafter\ifx\csname urlstyle\endcsname\relax
  \providecommand{\doi}[1]{doi: #1}\else
  \providecommand{\doi}{doi: \begingroup \urlstyle{rm}\Url}\fi

\bibitem[Adam et~al.(2025)Adam, {\v{C}}erm{\'a}k, Papafitsoros, and Picek]{adam2025wildlifereid}
Luk{\'a}{\v{s}} Adam, Vojt{\v{e}}ch {\v{C}}erm{\'a}k, Kostas Papafitsoros, and Lukas Picek.
\newblock Wildlifereid-10k: Wildlife re-identification dataset with 10k individual animals.
\newblock In \emph{Proceedings of the IEEE/CVF Conference on Computer Vision and Pattern Recognition workshops}, pages 2090--2100. IEEE, 2025.

\bibitem[Badger et~al.(2020)Badger, Wang, Modh, Perkes, Kolotouros, Pfrommer, Schmidt, and Daniilidis]{badger20203d}
Marc Badger, Yufu Wang, Adarsh Modh, Ammon Perkes, Nikos Kolotouros, Bernd~G Pfrommer, Marc~F Schmidt, and Kostas Daniilidis.
\newblock 3d bird reconstruction: a dataset, model, and shape recovery from a single view.
\newblock In \emph{European Conference on Computer Vision}, pages 1--17. Springer, 2020.

\bibitem[Bai et~al.(2025)Bai, Chen, Liu, Wang, Ge, Song, Dang, Wang, Wang, Tang, Zhong, Zhu, Yang, Li, Wan, Wang, Ding, Fu, Xu, Ye, Zhang, Xie, Cheng, Zhang, Yang, Xu, and Lin]{Qwen2.5-VL}
Shuai Bai, Keqin Chen, Xuejing Liu, Jialin Wang, Wenbin Ge, Sibo Song, Kai Dang, Peng Wang, Shijie Wang, Jun Tang, Humen Zhong, Yuanzhi Zhu, Mingkun Yang, Zhaohai Li, Jianqiang Wan, Pengfei Wang, Wei Ding, Zheren Fu, Yiheng Xu, Jiabo Ye, Xi Zhang, Tianbao Xie, Zesen Cheng, Hang Zhang, Zhibo Yang, Haiyang Xu, and Junyang Lin.
\newblock Qwen2.5-vl technical report.
\newblock \emph{arXiv preprint arXiv:2502.13923}, 2025.

\bibitem[Chen et~al.(2025)Chen, Guo, Zhu, Zhang, Huang, Feng, and Kang]{video_depth_anything}
Sili Chen, Hengkai Guo, Shengnan Zhu, Feihu Zhang, Zilong Huang, Jiashi Feng, and Bingyi Kang.
\newblock Video depth anything: Consistent depth estimation for super-long videos.
\newblock In \emph{Proceedings of the IEEE/CVF Computer Vision and Pattern Recognition Conference}, pages 22831--22840, 2025.

\bibitem[Chen et~al.(2024)Chen, Huang, Liu, Shen, Zhao, and Zhao]{chen2024anydoor}
Xi Chen, Lianghua Huang, Yu Liu, Yujun Shen, Deli Zhao, and Hengshuang Zhao.
\newblock Anydoor: Zero-shot object-level image customization.
\newblock In \emph{Proceedings of the IEEE/CVF Conference on Computer Vision and Pattern Recognition}, pages 6593--6602, 2024.

\bibitem[Chiquier et~al.(2025)Chiquier, Avrech, Gandelsman, Feng, Bouman, and Vondrick]{chiquier2025teaching}
Mia Chiquier, Orr Avrech, Yossi Gandelsman, Berthy Feng, Katherine Bouman, and Carl Vondrick.
\newblock Teaching humans subtle differences with diffusion.
\newblock \emph{arXiv preprint arXiv:2504.08046}, 2025.

\bibitem[Comanici et~al.(2025)Comanici, Bieber, Schaekermann, Pasupat, Sachdeva, Dhillon, Blistein, Ram, Zhang, Rosen, et~al.]{nanobanana}
Gheorghe Comanici, Eric Bieber, Mike Schaekermann, Ice Pasupat, Noveen Sachdeva, Inderjit Dhillon, Marcel Blistein, Ori Ram, Dan Zhang, Evan Rosen, et~al.
\newblock Gemini 2.5: Pushing the frontier with advanced reasoning, multimodality, long context, and next generation agentic capabilities.
\newblock \emph{arXiv preprint arXiv:2507.06261}, 2025.

\bibitem[Deitke et~al.(2023)Deitke, Liu, Wallingford, Ngo, Michel, Kusupati, Fan, Laforte, Voleti, Gadre, et~al.]{objaverseXL}
Matt Deitke, Ruoshi Liu, Matthew Wallingford, Huong Ngo, Oscar Michel, Aditya Kusupati, Alan Fan, Christian Laforte, Vikram Voleti, Samir~Yitzhak Gadre, et~al.
\newblock Objaverse-xl: A universe of 10m+ 3d objects.
\newblock \emph{Advances in Neural Information Processing Systems}, 36:\penalty0 35799--35813, 2023.

\bibitem[Ferreira et~al.(2020)Ferreira, Silva, Renna, Brandl, Renoult, Farine, Covas, and Doutrelant]{ferreira2020deep}
Andr{\'e}~C Ferreira, Liliana~R Silva, Francesco Renna, Hanja~B Brandl, Julien~P Renoult, Damien~R Farine, Rita Covas, and Claire Doutrelant.
\newblock Deep learning-based methods for individual recognition in small birds.
\newblock \emph{Methods in Ecology and Evolution}, 11\penalty0 (9):\penalty0 1072--1085, 2020.

\bibitem[Gal et~al.(2023)Gal, Alaluf, Atzmon, Patashnik, Bermano, Chechik, and Cohen-Or]{gal2022image}
Rinon Gal, Yuval Alaluf, Yuval Atzmon, Or Patashnik, Amit~Haim Bermano, Gal Chechik, and Daniel Cohen-Or.
\newblock An image is worth one word: Personalizing text-to-image generation using textual inversion.
\newblock In \emph{International Conference on Learning Representations}, 2023.

\bibitem[Ge et~al.(2019)Ge, Zhang, Wang, Tang, and Luo]{ge2019deepfashion2}
Yuying Ge, Ruimao Zhang, Xiaogang Wang, Xiaoou Tang, and Ping Luo.
\newblock Deepfashion2: A versatile benchmark for detection, pose estimation, segmentation and re-identification of clothing images.
\newblock In \emph{Proceedings of the IEEE/CVF Conference on Computer Vision and Pattern Recognition}, pages 5337--5345, 2019.

\bibitem[Ge et~al.(2016)Ge, McCool, Sanderson, Wang, Liu, Reid, and Corke]{ge2016temporal}
ZongYuan Ge, Chris McCool, Conrad Sanderson, Peng Wang, Lingqiao Liu, Ian Reid, and Peter Corke.
\newblock Exploiting temporal information for {DCNN}-based fine-grained object classification.
\newblock In \emph{International Conference on Digital Image Computing: Techniques and Applications}, 2016.

\bibitem[Goel et~al.(2020)Goel, Kanazawa, and Malik]{goel2020shape}
Shubham Goel, Angjoo Kanazawa, and Jitendra Malik.
\newblock Shape and viewpoint without keypoints.
\newblock In \emph{European Conference on Computer Vision}, pages 88--104. Springer, 2020.

\bibitem[H{\"a}gerlind et~al.(2024)H{\"a}gerlind, Hentati-Sundberg, and Wandt]{hagerlind2024temporally}
Johannes H{\"a}gerlind, Jonas Hentati-Sundberg, and Bastian Wandt.
\newblock Temporally-consistent 3d reconstruction of birds.
\newblock \emph{arXiv preprint arXiv:2408.13629}, 2024.

\bibitem[{iNaturalist}()]{iNaturalist}
{iNaturalist}.
\newblock \url{https://www.inaturalist.org}, 2025.
\newblock Accessed on Nov 10, 2025.

\bibitem[Joung et~al.(2021)Joung, Kim, Kim, Kim, and Sohn]{joung2021learning}
Sunghun Joung, Seungryong Kim, Minsu Kim, Ig-Jae Kim, and Kwanghoon Sohn.
\newblock Learning canonical 3d object representation for fine-grained recognition.
\newblock In \emph{Proceedings of the IEEE/CVF International Conference on Computer Vision}, pages 1035--1045, 2021.

\bibitem[Kanazawa et~al.(2018)Kanazawa, Tulsiani, Efros, and Malik]{kanazawa2018learning}
Angjoo Kanazawa, Shubham Tulsiani, Alexei~A Efros, and Jitendra Malik.
\newblock Learning category-specific mesh reconstruction from image collections.
\newblock In \emph{European Conference on Computer Vision}, pages 371--386, 2018.

\bibitem[Khurana et~al.(2024)Khurana, Daw, Maruf, Uyeda, Dahdul, Charpentier, Bak{\i}{\c{s}}, Bart~Jr, Mabee, Lapp, et~al.]{khurana2024hierarchical}
Mridul Khurana, Arka Daw, M Maruf, Josef~C Uyeda, Wasila Dahdul, Caleb Charpentier, Yasin Bak{\i}{\c{s}}, Henry~L Bart~Jr, Paula~M Mabee, Hilmar Lapp, et~al.
\newblock Hierarchical conditioning of diffusion models using tree-of-life for studying species evolution.
\newblock In \emph{European Conference on Computer Vision}, pages 137--153. Springer, 2024.

\bibitem[Kotar et~al.(2023)Kotar, Tian, Yu, Yamins, and Wu]{kotar2023these}
Klemen Kotar, Stephen Tian, Hong-Xing Yu, Dan Yamins, and Jiajun Wu.
\newblock Are these the same apple? comparing images based on object intrinsics.
\newblock \emph{Advances in Neural Information Processing Systems}, 36:\penalty0 40853--40871, 2023.

\bibitem[Labs et~al.(2025)Labs, Batifol, Blattmann, Boesel, Consul, Diagne, Dockhorn, English, English, Esser, et~al.]{labs2025flux}
Black~Forest Labs, Stephen Batifol, Andreas Blattmann, Frederic Boesel, Saksham Consul, Cyril Diagne, Tim Dockhorn, Jack English, Zion English, Patrick Esser, et~al.
\newblock Flux. 1 kontext: Flow matching for in-context image generation and editing in latent space.
\newblock \emph{arXiv preprint arXiv:2506.15742}, 2025.

\bibitem[Li et~al.(2020)Li, Liu, Kim, De~Mello, Jampani, Yang, and Kautz]{li2020self}
Xueting Li, Sifei Liu, Kihwan Kim, Shalini De~Mello, Varun Jampani, Ming-Hsuan Yang, and Jan Kautz.
\newblock Self-supervised single-view 3d reconstruction via semantic consistency.
\newblock In \emph{European Conference on Computer Vision}, pages 677--693. Springer, 2020.

\bibitem[Li et~al.(2021)Li, Takehara, Taketomi, Zheng, and Nie{\ss}ner]{li20214dcomplete}
Yang Li, Hikari Takehara, Takafumi Taketomi, Bo Zheng, and Matthias Nie{\ss}ner.
\newblock 4dcomplete: Non-rigid motion estimation beyond the observable surface.
\newblock In \emph{Proceedings of the IEEE/CVF International Conference on Computer Vision}, pages 12706--12716, 2021.

\bibitem[Mishchenko and Defazio(2024)]{mishchenko2024prodigy}
Konstantin Mishchenko and Aaron Defazio.
\newblock Prodigy: An expeditiously adaptive parameter-free learner.
\newblock In \emph{International Conference on Machine Learning}, pages 35779--35804. PMLR, 2024.

\bibitem[Monsefi et~al.(2025)Monsefi, Khurana, Ramnath, Karpatne, Chao, and Zhang]{monsefi2025taxadiffusion}
Amin~Karimi Monsefi, Mridul Khurana, Rajiv Ramnath, Anuj Karpatne, Wei-Lun Chao, and Cheng Zhang.
\newblock Taxadiffusion: Progressively trained diffusion model for fine-grained species generation.
\newblock In \emph{Proceedings of the IEEE/CVF International Conference on Computer Vision}, pages 8579--8589, 2025.

\bibitem[Moradi et~al.(2025)Moradi, Haque, Kaur, Bentz, Bridge, and Habibi]{moradi2025context}
Keon Moradi, Ethan Haque, Jasmeen Kaur, Alexandra~B Bentz, Eli~S Bridge, and Golnaz Habibi.
\newblock Context-aware outlier rejection for robust multi-view 3d tracking of similar small birds in an outdoor aviary.
\newblock In \emph{Proceedings of the IEEE/CVF Winter Conference on Applications of Computer Vision}, pages 983--991. IEEE, 2025.

\bibitem[Mou et~al.(2025)Mou, Wu, Wu, Guo, Zhang, Cheng, Luo, Ding, Zhang, Li, et~al.]{mou2025dreamo}
Chong Mou, Yanze Wu, Wenxu Wu, Zinan Guo, Pengze Zhang, Yufeng Cheng, Yiming Luo, Fei Ding, Shiwen Zhang, Xinghui Li, et~al.
\newblock Dreamo: A unified framework for image customization.
\newblock In \emph{Proceedings of the SIGGRAPH Asia 2025 Conference Papers}, pages 1--12, 2025.

\bibitem[Naik et~al.(2023)Naik, Chan, Yang, Delacoux, Couzin, Kano, and Nagy]{naik20233d}
Hemal Naik, Alex Hoi~Hang Chan, Junran Yang, Mathilde Delacoux, Iain~D Couzin, Fumihiro Kano, and M{\'a}t{\'e} Nagy.
\newblock 3d-pop-an automated annotation approach to facilitate markerless 2d-3d tracking of freely moving birds with marker-based motion capture.
\newblock In \emph{Proceedings of the IEEE/CVF Conference on Computer Vision and Pattern Recognition}, pages 21274--21284, 2023.

\bibitem[Ng et~al.(2025)Ng, Yang, Sii, Deng, Chan, Song, Xiang, and Zhu]{ng2024chirpy3d}
Kam~Woh Ng, Jing Yang, Jia~Wei Sii, Jiankang Deng, Chee~Seng Chan, Yi-Zhe Song, Tao Xiang, and Xiatian Zhu.
\newblock Chirpy3d: Creative fine-grained 3d object fabrication via part sampling.
\newblock \emph{arXiv preprint arXiv:2501.04144}, 2025.

\bibitem[OpenAI(2023)]{gpt4v}
OpenAI.
\newblock Gpt-4v(ision) system card, 2023.

\bibitem[Oquab et~al.(2024)Oquab, Darcet, Moutakanni, Vo, Szafraniec, Khalidov, Fernandez, Haziza, Massa, El-Nouby, et~al.]{oquab2023dinov2}
Maxime Oquab, Timoth{\'e}e Darcet, Th{\'e}o Moutakanni, Huy Vo, Marc Szafraniec, Vasil Khalidov, Pierre Fernandez, Daniel Haziza, Francisco Massa, Alaaeldin El-Nouby, et~al.
\newblock Dinov2: Learning robust visual features without supervision.
\newblock \emph{Transactions on Machine Learning Research Journal}, 2024.

\bibitem[Pan et~al.(2025)Pan, Wang, Li, He, and Lai]{pan2025finediffusion}
Ziying Pan, Kun Wang, Gang Li, Feihong He, and Yongxuan Lai.
\newblock Finediffusion: scaling up diffusion models for fine-grained image generation with 10,000 classes.
\newblock \emph{Applied Intelligence}, 55\penalty0 (5):\penalty0 309, 2025.

\bibitem[Peebles and Xie(2023)]{peebles2023scalable}
William Peebles and Saining Xie.
\newblock Scalable diffusion models with transformers.
\newblock In \emph{Proceedings of the IEEE/CVF International Conference on Computer Vision}, pages 4195--4205, 2023.

\bibitem[Peng et~al.(2025)Peng, Cui, Tang, Qi, Dong, Bai, Han, Ge, Zhang, and Xia]{peng2024dreambench}
Yuang Peng, Yuxin Cui, Haomiao Tang, Zekun Qi, Runpei Dong, Jing Bai, Chunrui Han, Zheng Ge, Xiangyu Zhang, and Shu-Tao Xia.
\newblock Dreambench++: A human-aligned benchmark for personalized image generation.
\newblock In \emph{International Conference on Learning Representations}, 2025.

\bibitem[Ravi et~al.(2024)Ravi, Gabeur, Hu, Hu, Ryali, Ma, Khedr, R{\"a}dle, Rolland, Gustafson, et~al.]{ravi2024sam2}
Nikhila Ravi, Valentin Gabeur, Yuan-Ting Hu, Ronghang Hu, Chaitanya Ryali, Tengyu Ma, Haitham Khedr, Roman R{\"a}dle, Chloe Rolland, Laura Gustafson, et~al.
\newblock Sam 2: Segment anything in images and videos.
\newblock In \emph{International Conference on Learning Representations}, 2024.

\bibitem[Ren et~al.(2024)Ren, Liu, Zeng, Lin, Li, Cao, Chen, Huang, Chen, Yan, et~al.]{ren2024grounded}
Tianhe Ren, Shilong Liu, Ailing Zeng, Jing Lin, Kunchang Li, He Cao, Jiayu Chen, Xinyu Huang, Yukang Chen, Feng Yan, et~al.
\newblock Grounded sam: Assembling open-world models for diverse visual tasks.
\newblock \emph{arXiv preprint arXiv:2401.14159}, 2024.

\bibitem[Rodriguez-Juan et~al.(2025)Rodriguez-Juan, Ortiz-Perez, Benavent-Lledo, Mulero-P{\'e}rez, Ruiz-Ponce, Orihuela-Torres, Garcia-Rodriguez, and Sebasti{\'a}n-Gonz{\'a}lez]{rodriguez2025visual}
Javier Rodriguez-Juan, David Ortiz-Perez, Manuel Benavent-Lledo, David Mulero-P{\'e}rez, Pablo Ruiz-Ponce, Adrian Orihuela-Torres, Jose Garcia-Rodriguez, and Esther Sebasti{\'a}n-Gonz{\'a}lez.
\newblock Visual wetlandbirds dataset: Bird species identification and behavior recognition in videos.
\newblock \emph{Scientific Data}, 12\penalty0 (1):\penalty0 1200, 2025.

\bibitem[Rombach et~al.(2022)Rombach, Blattmann, Lorenz, Esser, and Ommer]{rombach2022high}
Robin Rombach, Andreas Blattmann, Dominik Lorenz, Patrick Esser, and Bj{\"o}rn Ommer.
\newblock High-resolution image synthesis with latent diffusion models.
\newblock In \emph{Proceedings of the IEEE/CVF Conference on Computer Vision and Pattern Recognition}, pages 10684--10695, 2022.

\bibitem[Ruiz et~al.(2023)Ruiz, Li, Jampani, Pritch, Rubinstein, and Aberman]{ruiz2023dreambooth}
Nataniel Ruiz, Yuanzhen Li, Varun Jampani, Yael Pritch, Michael Rubinstein, and Kfir Aberman.
\newblock Dreambooth: Fine tuning text-to-image diffusion models for subject-driven generation.
\newblock In \emph{Proceedings of the IEEE/CVF Conference on Computer Vision and Pattern Recognition}, pages 22500--22510, 2023.

\bibitem[Saha et~al.(2024)Saha, Van~Horn, and Maji]{saha2024improved}
Oindrila Saha, Grant Van~Horn, and Subhransu Maji.
\newblock Improved zero-shot classification by adapting vlms with text descriptions.
\newblock In \emph{Proceedings of the IEEE/CVF Conference on Computer Vision and Pattern Recognition}, pages 17542--17552, 2024.

\bibitem[Saha et~al.(2025)Saha, Lawrence, Van~Horn, and Maji]{saha2025generate}
Oindrila Saha, Logan Lawrence, Grant Van~Horn, and Subhransu Maji.
\newblock Generate, transduct, adapt: Iterative transduction with vlms.
\newblock In \emph{Proceedings of the IEEE/CVF International Conference on Computer Vision}, pages 1369--1379, 2025.

\bibitem[Saito et~al.(2016)Saito, Kanezaki, and Harada]{7552915}
Tomoaki Saito, Asako Kanezaki, and Tatsuya Harada.
\newblock Ibc127: Video dataset for fine-grained bird classification.
\newblock In \emph{2016 IEEE International Conference on Multimedia and Expo}, pages 1--6, 2016.

\bibitem[Shinoda and Shiohara(2024)]{shinoda2024petface}
Risa Shinoda and Kaede Shiohara.
\newblock Petface: A large-scale dataset and benchmark for animal identification.
\newblock In \emph{European Conference on Computer Vision}, pages 19--36. Springer, 2024.

\bibitem[Song et~al.(2026)Song, Jiang, Yang, Cheng, Quan, and Yang]{song2025insert}
Wensong Song, Hong Jiang, Zongxin Yang, Zheqiao Cheng, Ruijie Quan, and Yi Yang.
\newblock Insert anything: Image insertion via in-context editing in dit.
\newblock In \emph{Proceedings of the AAAI Conference on Artificial Intelligence}, pages 9097--9105, 2026.

\bibitem[Sullivan et~al.(2009)Sullivan, Wood, Iliff, Bonney, Fink, and Kelling]{sullivan2009ebird}
Brian~L Sullivan, Christopher~L Wood, Marshall~J Iliff, Rick~E Bonney, Daniel Fink, and Steve Kelling.
\newblock ebird: A citizen-based bird observation network in the biological sciences.
\newblock \emph{Biological conservation}, 142\penalty0 (10):\penalty0 2282--2292, 2009.

\bibitem[Sun et~al.(2022)Sun, Wang, Cai, Wang, Huang, Li, Shao, and Wang]{Sun_2022_ACCV}
Hongyu Sun, Yongcai Wang, Xudong Cai, Peng Wang, Zhe Huang, Deying Li, Yu Shao, and Shuo Wang.
\newblock Airbirds: A large-scale challenging dataset for bird strike prevention in real-world airports.
\newblock In \emph{Proceedings of the Asian Conference on Computer Vision}, pages 2440--2456, 2022.

\bibitem[Sun et~al.(2025)Sun, Hua, Li, Qi, Li, Li, and Zhang]{sun2024fbdsv2024}
Zi-Wei Sun, Ze-Xi Hua, Heng-Chao Li, Zhi-Peng Qi, Xiang Li, Yan Li, and Jin-Chi Zhang.
\newblock Fbd-sv-2024: Flying bird object detection dataset in surveillance video.
\newblock \emph{Scientific Data}, 12\penalty0 (1):\penalty0 530, 2025.

\bibitem[Sundaram et~al.(2025)Sundaram, Chae, Tian, Beery, and Isola]{sundarampersonalized}
Shobhita Sundaram, Julia Chae, Yonglong Tian, Sara Beery, and Phillip Isola.
\newblock Personalized representation from personalized generation.
\newblock In \emph{International Conference on Learning Representations}, 2025.

\bibitem[Tan et~al.(2025)Tan, Liu, Yang, Xue, and Wang]{tan2025ominicontrol}
Zhenxiong Tan, Songhua Liu, Xingyi Yang, Qiaochu Xue, and Xinchao Wang.
\newblock Ominicontrol: Minimal and universal control for diffusion transformer.
\newblock In \emph{Proceedings of the IEEE/CVF International Conference on Computer Vision}, pages 14940--14950, 2025.

\bibitem[Van~Horn et~al.(2015)Van~Horn, Branson, Farrell, Haber, Barry, Ipeirotis, Perona, and Belongie]{van2015building}
Grant Van~Horn, Steve Branson, Ryan Farrell, Scott Haber, Jessie Barry, Panos Ipeirotis, Pietro Perona, and Serge Belongie.
\newblock Building a bird recognition app and large scale dataset with citizen scientists: The fine print in fine-grained dataset collection.
\newblock In \emph{Proceedings of the IEEE/CVF Conference on Computer Vision and Pattern Recognition}, pages 595--604, 2015.

\bibitem[Van~Horn et~al.(2022)Van~Horn, Qian, Wilber, Adam, Mac~Aodha, and Belongie]{ssw602022eccv}
Grant Van~Horn, Rui Qian, Kimberly Wilber, Hartwig Adam, Oisin Mac~Aodha, and Serge Belongie.
\newblock Exploring fine-grained audiovisual categorization with the ssw60 dataset.
\newblock In \emph{European Conference on Computer Vision}, pages 271--289. Springer, 2022.

\bibitem[Wah et~al.(2011)Wah, Branson, Welinder, Perona, and Belongie]{WahCUB_200_2011}
C. Wah, S. Branson, P. Welinder, P. Perona, and S. Belongie.
\newblock The caltech-ucsd birds-200-2011 dataset.
\newblock Technical Report CNS-TR-2011-001, California Institute of Technology, 2011.

\bibitem[Wang et~al.(2023)Wang, Que, Chen, Li, Li, and Yang]{wang2023creative}
Renke Wang, Guimin Que, Shuo Chen, Xiang Li, Jun Li, and Jian Yang.
\newblock Creative birds: self-supervised single-view 3d style transfer.
\newblock In \emph{Proceedings of the IEEE/CVF International Conference on Computer Vision}, pages 8775--8784, 2023.

\bibitem[Wang et~al.(2021)Wang, Kolotouros, Daniilidis, and Badger]{wang2021birds}
Yufu Wang, Nikos Kolotouros, Kostas Daniilidis, and Marc Badger.
\newblock Birds of a feather: Capturing avian shape models from images.
\newblock In \emph{Proceedings of the IEEE/CVF Conference on Computer Vision and Pattern Recognition}, pages 14739--14749, 2021.

\bibitem[Wu et~al.(2023)Wu, Li, Jakab, Rupprecht, and Vedaldi]{wu2023magicpony}
Shangzhe Wu, Ruining Li, Tomas Jakab, Christian Rupprecht, and Andrea Vedaldi.
\newblock Magicpony: Learning articulated 3d animals in the wild.
\newblock In \emph{Proceedings of the IEEE/CVF Conference on Computer Vision and Pattern Recognition}, pages 8792--8802, 2023.

\bibitem[Xu et~al.(2023)Xu, Zhang, Peng, Ma, Jesslen, Ji, Hu, Zhang, Liu, Wang, et~al.]{xu2023animal3d}
Jiacong Xu, Yi Zhang, Jiawei Peng, Wufei Ma, Artur Jesslen, Pengliang Ji, Qixin Hu, Jiehua Zhang, Qihao Liu, Jiahao Wang, et~al.
\newblock Animal3d: A comprehensive dataset of 3d animal pose and shape.
\newblock In \emph{Proceedings of the IEEE/CVF International Conference on Computer Vision}, pages 9099--9109, 2023.

\bibitem[Zhai et~al.(2023)Zhai, Mustafa, Kolesnikov, and Beyer]{zhai2023sigmoid}
Xiaohua Zhai, Basil Mustafa, Alexander Kolesnikov, and Lucas Beyer.
\newblock Sigmoid loss for language image pre-training.
\newblock In \emph{Proceedings of the IEEE/CVF International Conference on Computer Vision}, pages 11975--11986, 2023.

\bibitem[Zhang et~al.(2023)Zhang, Rao, and Agrawala]{zhang2023adding}
Lvmin Zhang, Anyi Rao, and Maneesh Agrawala.
\newblock Adding conditional control to text-to-image diffusion models.
\newblock In \emph{Proceedings of the IEEE/CVF International Conference on Computer Vision}, pages 3836--3847, 2023.

\bibitem[Zhang et~al.(2018)Zhang, Isola, Efros, Shechtman, and Wang]{zhang2018perceptual}
Richard Zhang, Phillip Isola, Alexei~A Efros, Eli Shechtman, and Oliver Wang.
\newblock The unreasonable effectiveness of deep features as a perceptual metric.
\newblock In \emph{Proceedings of the IEEE/CVF Conference on Computer Vision and Pattern Recognition}, pages 586--595, 2018.

\bibitem[Zuffi et~al.(2017)Zuffi, Kanazawa, Jacobs, and Black]{Zuffi:CVPR:2017}
Silvia Zuffi, Angjoo Kanazawa, David~W Jacobs, and Michael~J Black.
\newblock 3d menagerie: Modeling the 3d shape and pose of animals.
\newblock In \emph{Proceedings of the IEEE conference on computer vision and pattern recognition}, pages 6365--6373, 2017.

\end{thebibliography}
}

\clearpage
\setcounter{page}{1}
\maketitlesupplementary
\appendix
\setcounter{table}{0}
\renewcommand{\thetable}{A\arabic{table}}
\setcounter{figure}{0}
\renewcommand{\thefigure}{A\arabic{figure}}

\section{Additional Dataset Information}
\label{sec:supp_datasets}
\subsection{\datasetname Annotation}
\label{sec:annotation}
\begin{figure*}[h!]
    \centering
    \includegraphics[width=0.8\textwidth]{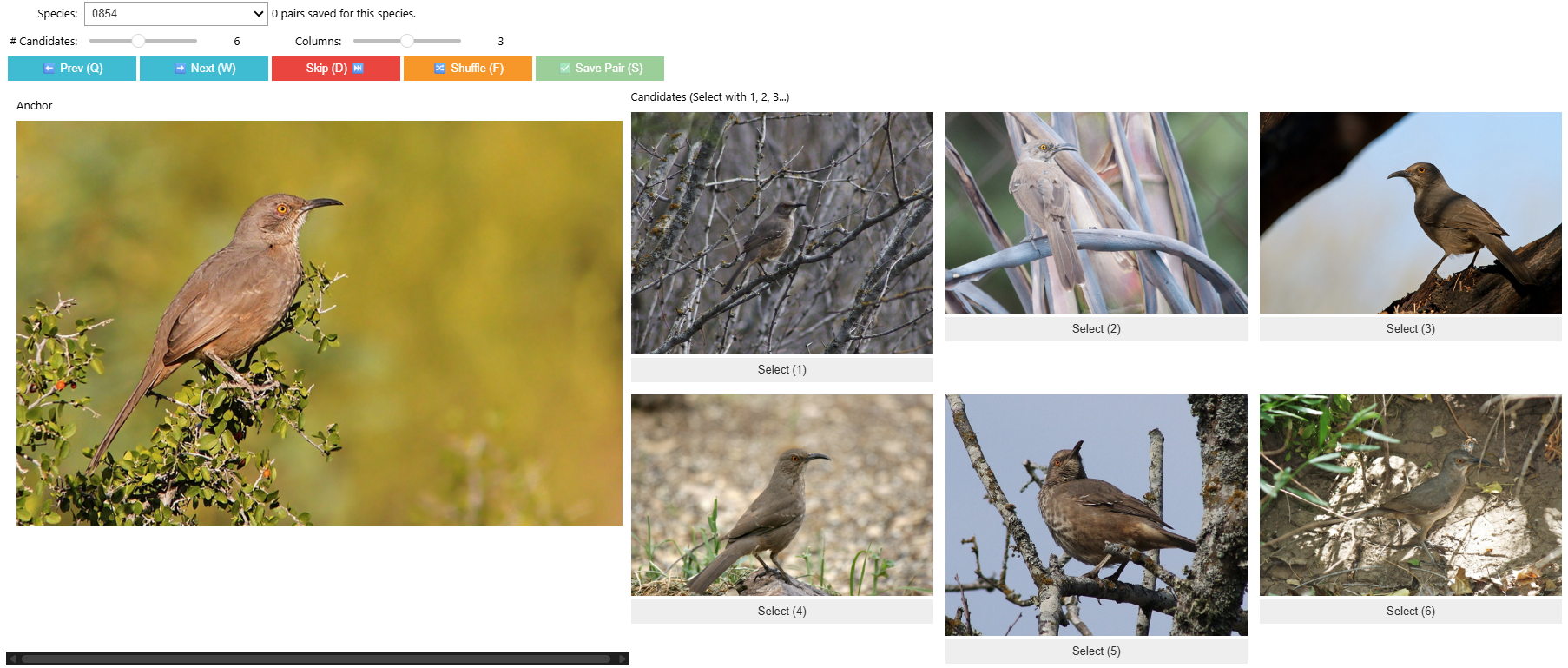}
    \caption{
 \textbf{\datasetname annotation widget. }
    Annotators are given the task of selecting an image where the individual present in the image looks the same as the individual in the anchor image. 
    By default, 6 candidate images are shown at random from the same class.
    Annotators have the option to shuffle the candidates or skip the anchor image.
    }
    \label{fig:supp_annotation}
\end{figure*}

A small group of birdwatcher volunteers with 8+ years of active field experience annotated the \datasetname dataset. 
For each datapoint, a subject image was selected at random and then 6 images of the same class were displayed to the annotator (Figure~\ref{fig:supp_annotation}). 
The annotator was given the task to select the image where the individual appears the most similar to the subject image. 
More specifically, we asked the annotators to evaluate the similarity of the birds in the two images using the following criteria:
\begin{enumerate}
    \item Plumage: Do the color patterns and textures on the surfaces of the birds appear the same? 
    \item Structure: Do the body shapes and proportions (\eg bill length, tail length, head shape, \etc) of the two birds appear the same?
\end{enumerate}
We asked annotators to choose images which contained individuals with matching plumage and structure such that they could plausibly be the same bird in a different setting.
Some examples of look-alike and non-look-alike pairs are given in Figure~\ref{fig:supp_mismatch}, along with reasons for why the negative pairs would not be considered a match.
Note these criteria do not include matching bird pose or background lighting conditions, since we explicitly wanted a diversity of conditions for generation.
If none of the candidate images were close enough, the annotator was given the option to shuffle the candidate images or to skip the subject image. 
This was repeated until approximately 5-10 images were selected for each class.

\subsection{Dataset Statistics}
\label{sec:supp_dataset_stats}
\begin{table}
    \centering
    \begin{tabular}{c | c | c | c}
        \hline\hline
         Dataset & \# Pairs & \# Classes & \# Species \\
        \hline
        \datasetname & 4759 & 539 & 401 \\
        iNat-Seen & 677 & 395 & 395 \\
        iNat-Unseen & 396 & 396 & 396 \\
        \hline\hline
    \end{tabular}
    \caption{
    \textbf{Dataset statistics for \datasetname, iNat-Seen, and iNat-Unseen.}
    Each pair contains two unique images, meaning the number of images is just twice the number of pairs.
    }
    \label{tab:supp_dataset_stats}
\end{table}

\begin{figure}[h!]
    \centering
    \includegraphics[width=0.25\textwidth]{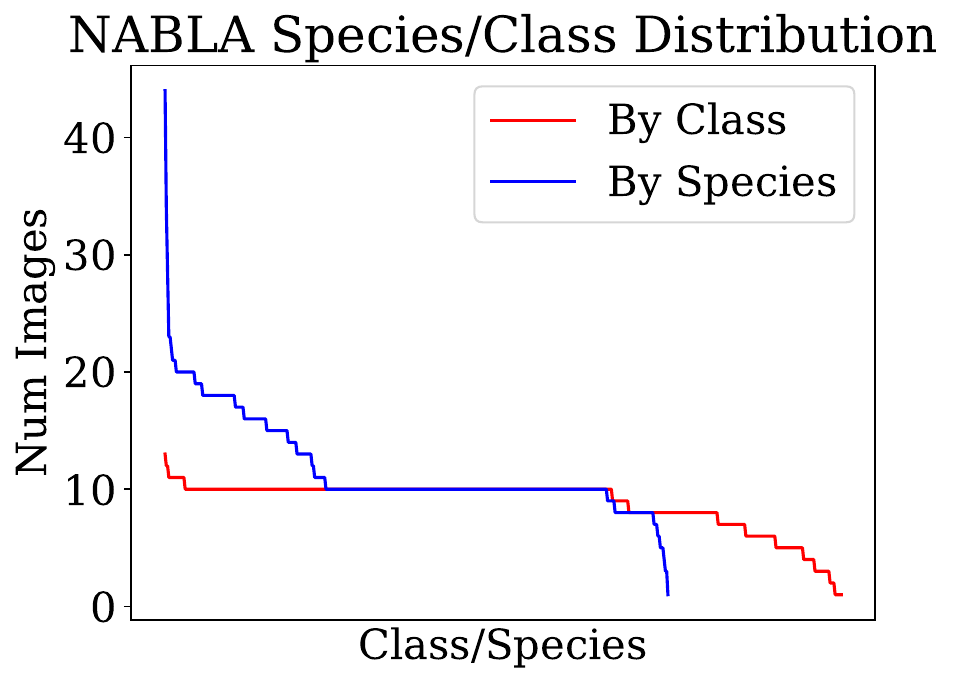}
    \includegraphics[width=0.21\textwidth]{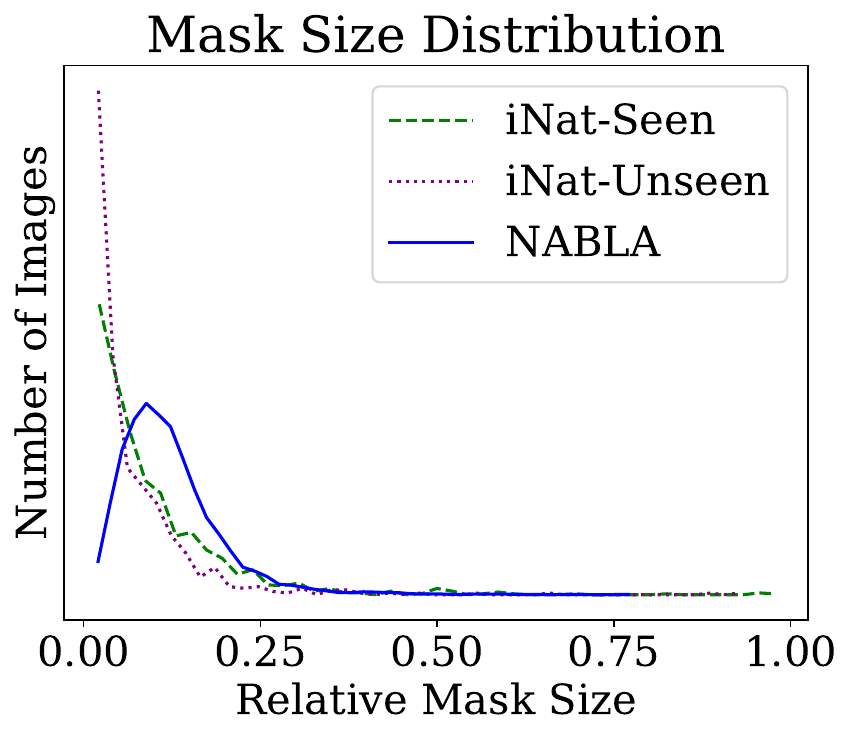}
    \caption{
    \textbf{\datasetname, iNat-Seen, and iNat-Unseen statistics.}
    Left: Most classes in \datasetname have 10 pairs.
    Right: Images in \datasetname typically have larger subjects than iNat-Seen or iNat-Unseen images. 
    }
    \label{fig:supp_dataset_graphs}
\end{figure}

\begin{figure}[h!]
    \centering
    \includegraphics[width=0.47\textwidth]{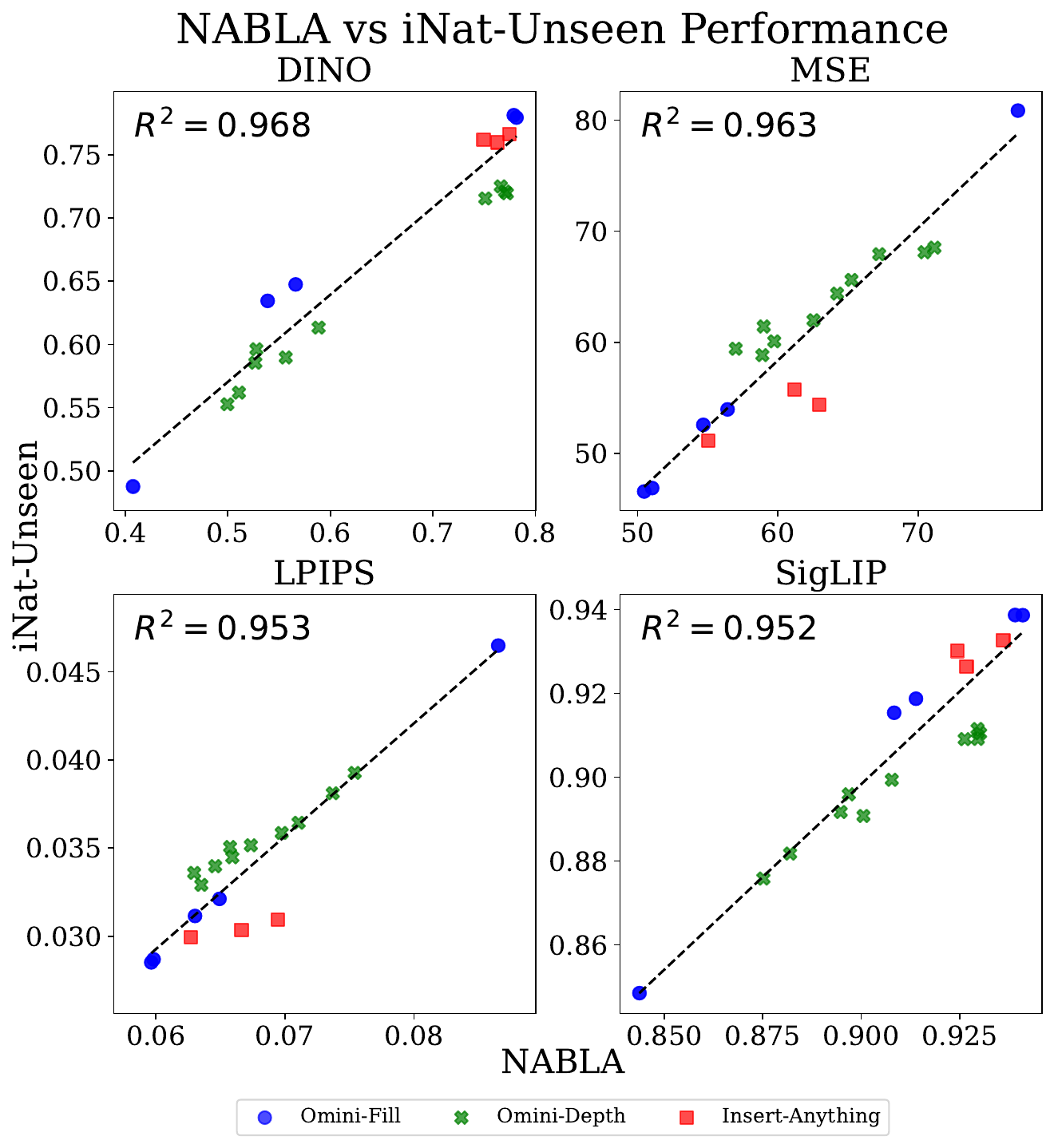}
    \caption{
    \textbf{NABLA and iNat-Unseen Performance Correlation.}
    Comparison of average model performance on \datasetname and iNat-Unseen data across all trials.
    }
    \label{fig:supp_graph_nabla_unseen}
\end{figure}
\begin{figure}[h!]
    \centering
    \includegraphics[width=0.47\textwidth]{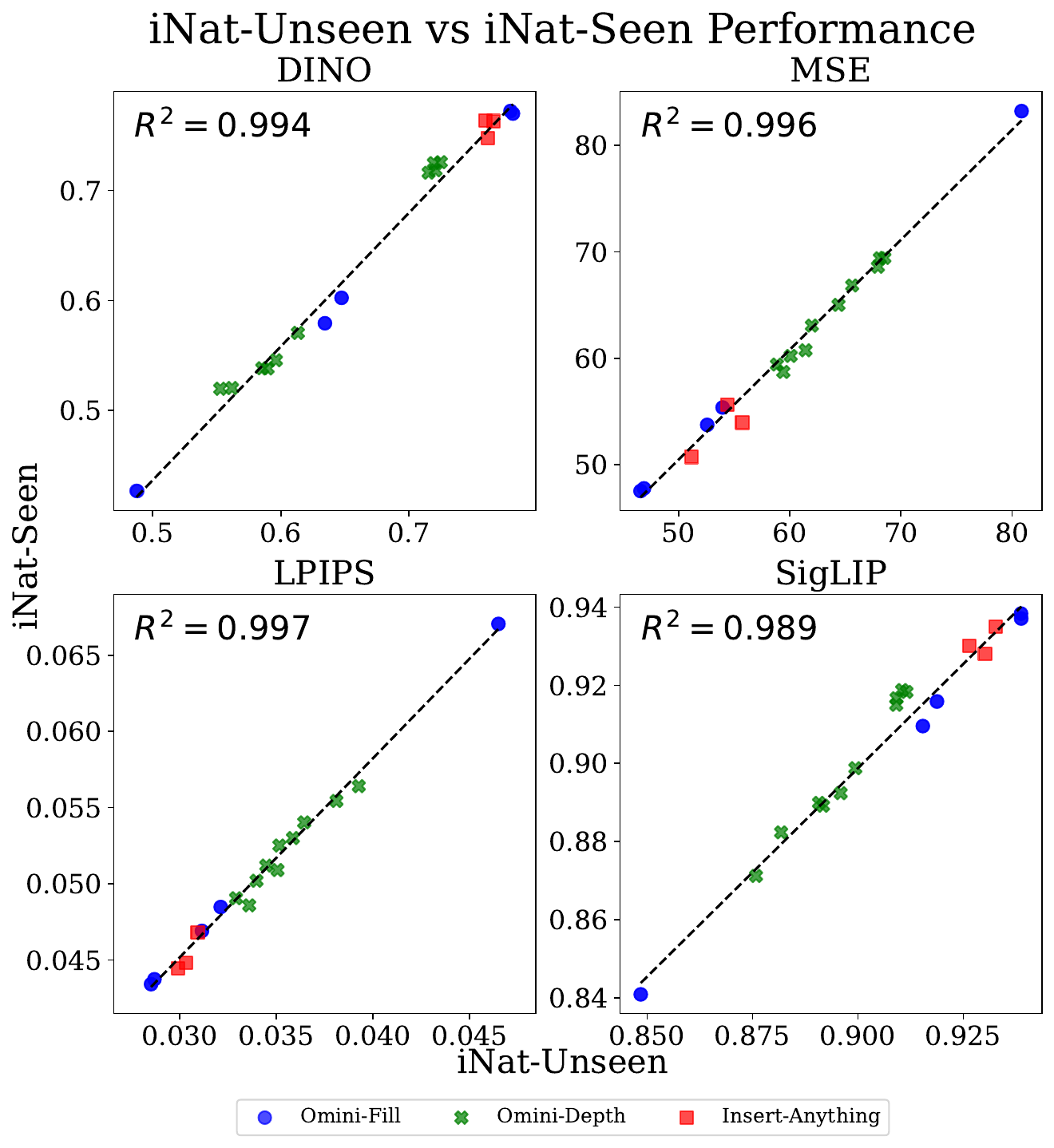}
    \caption{
    \textbf{iNat-Unseen and iNat-Seen Performance Correlation.}
    Comparison of average model performance on iNat-Unseen and iNat-Seen data across all trials.
    }
    \label{fig:supp_graph_inat_both}
\end{figure}

In Table~\ref{tab:supp_dataset_stats} we outline the basic statistics for the \datasetname, iNat-Seen, and iNat-Unseen datasets.
We see that \datasetname contains 539 classes across 401 species, meaning over 100 species have different classes based on age, sex, or breeding status.
Furthermore, a majority of classes in \datasetname have at least 10 image pairs, shown in the left plot of Figure~\ref{fig:supp_dataset_graphs}. 

To create the iNaturalist datasets, we queried the iNaturalist API to get recent observations based on species name for research quality data.
For iNat-Seen, we used the species list of NABirds and sampled 1-2 observations per species. 
For iNat-Unseen, we used the Aves species list from the 2017 iNaturalist competition, excluding the birds in NABirds, and sampled 1 observation per species. 
For observations with more than 2 images, we randomly selected 2 images to serve as the representative pair. 

In the right plot of Figure~\ref{fig:supp_dataset_graphs} we see NABLA images generally have larger subjects than iNaturalist data. 
We consider this to be correlated with image quality, indicating NABLA images to generally be higher quality than our iNat datasets.

\begin{table}
    \centering
    \begin{tabular}{c | c | c | c | c}
        \hline\hline
        Control & Backbone & Train & Test & DINO \\
        \hline
        Depth & Kontext & Short & Short & 0.77 \\
        Depth & Kontext & Short & Long & 0.75 \\
        Depth & Kontext & Long & Short & 0.77 \\
        Depth & Kontext & Long & Long & \textbf{0.77} \\
        \hline
        Depth & Schnell & Short & Short & 0.53 \\
        Depth & Schnell & Short & Long & 0.53 \\
        Depth & Schnell & Long & Short & 0.56 \\
        Depth & Schnell & Long & Long & \textbf{0.59} \\
        \hline\hline
    \end{tabular}
    \caption{
    \textbf{Caption length ablation.} Training and evaluating on the long captions works better. 
    }
    \label{tab:supp_caption_results}
\end{table}

\subsection{Caption Generation}
\label{sec:supp_caption}
We explore two different captioning modes for training and evaluation. 
In both cases we use Qwen-2.5 VL for generating captions on each image, but vary the prompt to get short captions and long captions. 
The prompt for each setting is given in Table~\ref{tab:supp_caption} along with examples for each one. 
We found in Table~\ref{tab:supp_caption_results} the long captions worked better but the differences were quite small overall. 

\begin{table*}
    \begin{tabular}{p{0.2\linewidth} | p{0.34\linewidth}| p{0.34\linewidth}}
        \hline\hline
         & Short Caption & Long Caption \\
         \hline
        Prompt & Look at this image and describe where the SUBJECT\_ITEM : `bird' is placed. Be *very* brief but do not miss elements EXCEPT the SUBJECT\_ITEM. DO NOT DESCRIBE OR MENTION THE SUBJECT\_ITEM. You should output starting with: ``Place it in'' or + ``Place it on.'' & Look at this image and describe where the bird is and what the background of the image is. DO NOT DESCRIBE OR MENTION THE POSE OR APPEARANCE OF THE BIRD.\\
        \hline 
        \includegraphics[width=\linewidth]{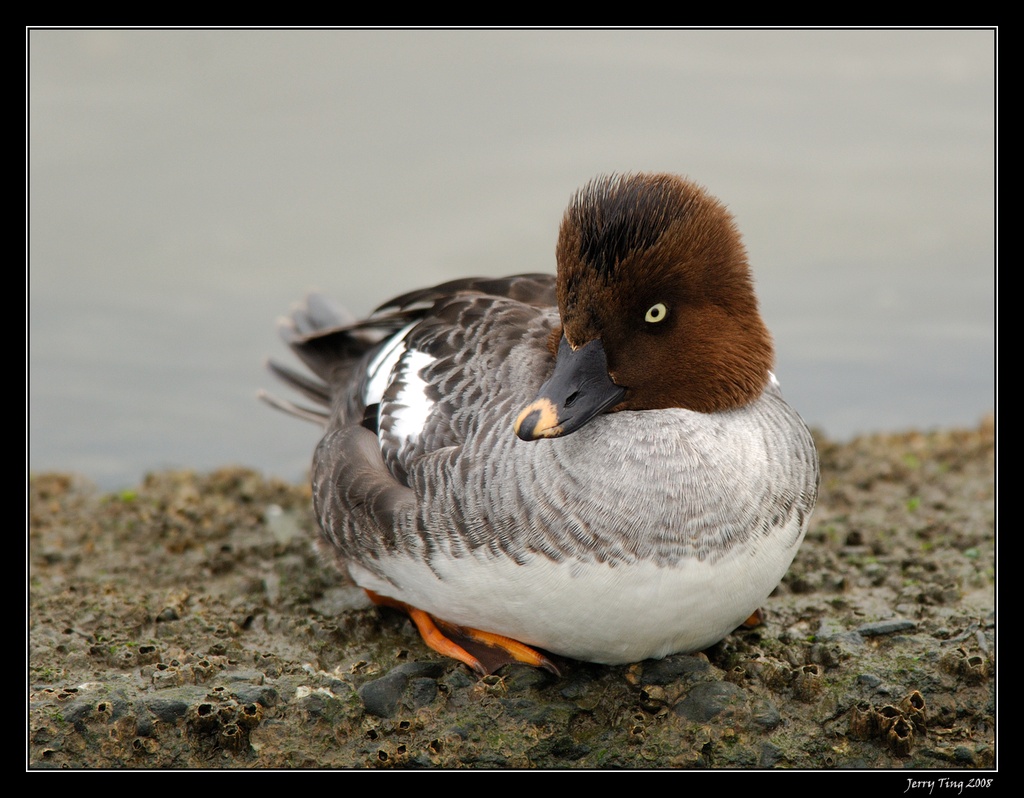} & ``Place it on a rocky surface near water.'' & ``The bird is on a rocky surface near a body of water, which appears to be calm and overcast. The background consists of the water and the sky, which is gray and cloudy.''\\
        \hline
        \includegraphics[width=\linewidth]{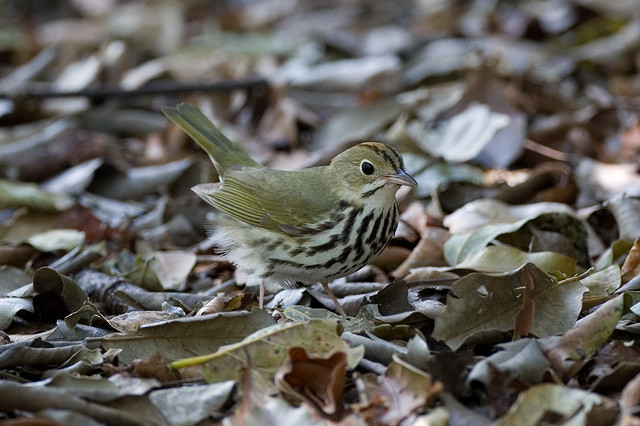} & ``Place it on leaves.'' & ``The bird is standing on the ground, surrounded by a bed of dried leaves. The background consists of scattered leaves in various shades of brown and green, creating a natural forest floor setting.''\\
        \hline\hline
    \end{tabular}
    \caption{
    \textbf{Sample captions from \datasetname generated using Qwen2.5 VL.}
    Each image has two captions, one long and one short. 
    We tried training and evaluating the depth and keypoint models using short and long captions but generally found better results using the long caption.
    }
    \label{tab:supp_caption}
\end{table*}
\begin{figure}[h!]
    \centering
    \includegraphics[width=0.45\textwidth]{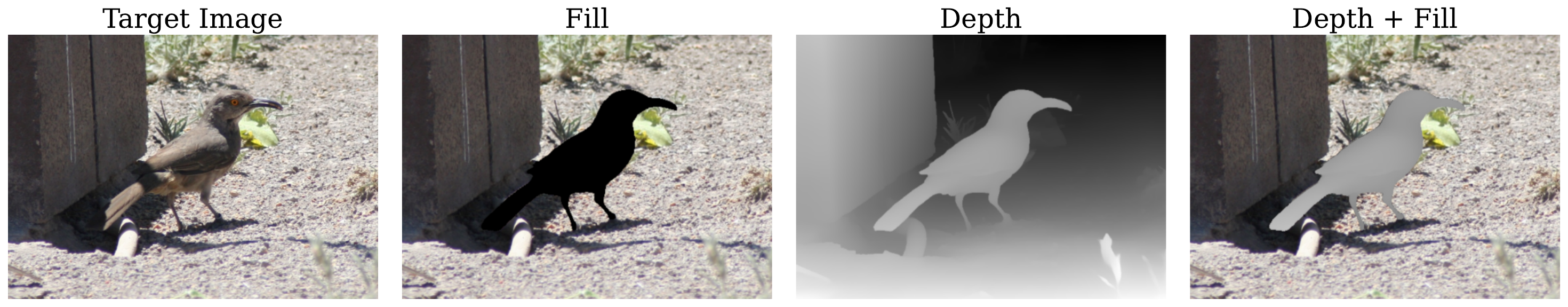}
    \caption{
 \textbf{Mixed depth + fill control.}
    The mixed control mode is a simple combination of the depth map and background fill image, where the mask is replaced with the pixels from the depth map. 
    }
    \label{fig:supp_mixed}
\end{figure}

\subsection{Additional Visualizations}
\label{sec:supp_additional_visualizations}
\begin{figure}[h!]
    \centering
    \datasetname Additional Examples\par\medskip
    \includegraphics[width=0.45\textwidth]{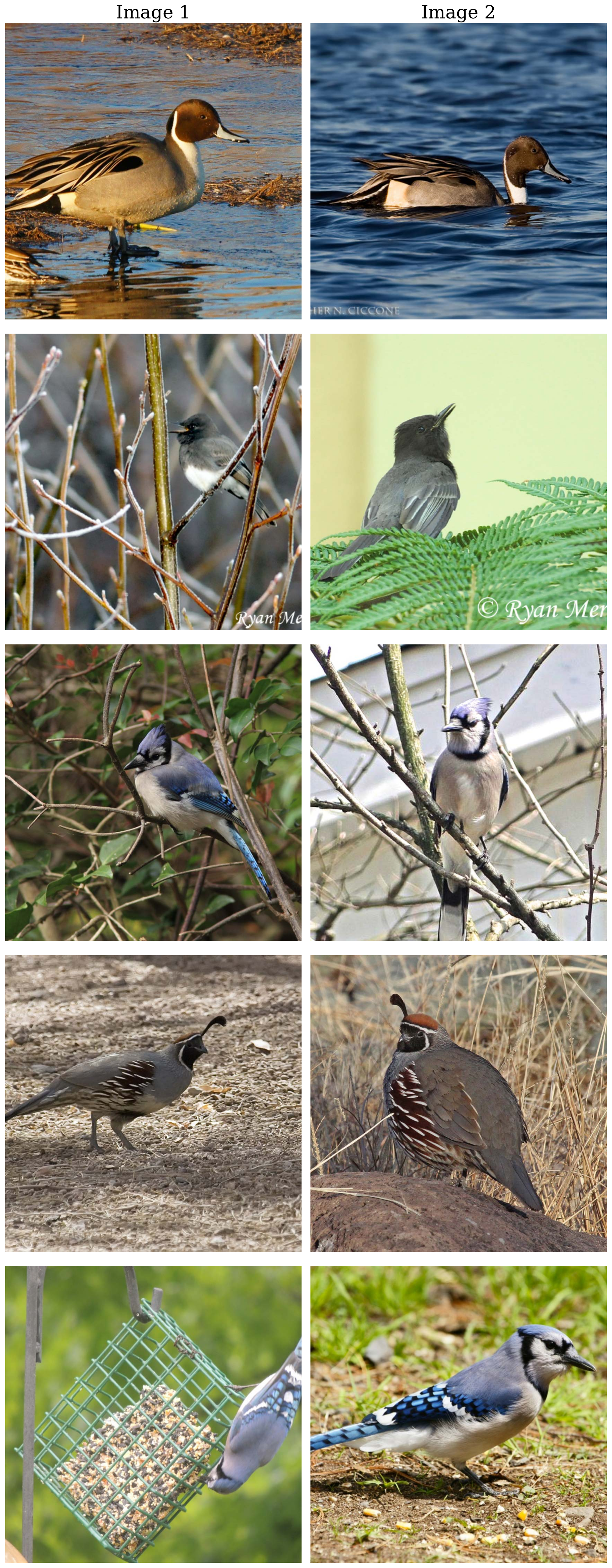}
    \caption{
        \textbf{Additional \datasetname examples.}
        Additional test examples from \datasetname, sampled randomly. 
        Images center-cropped to square.
    }
    \label{fig:supp_nabla}
\end{figure}

\begin{figure}[h!]
    \centering
    iNat-Seen Additional Examples\par\medskip
    \includegraphics[width=0.45\textwidth]{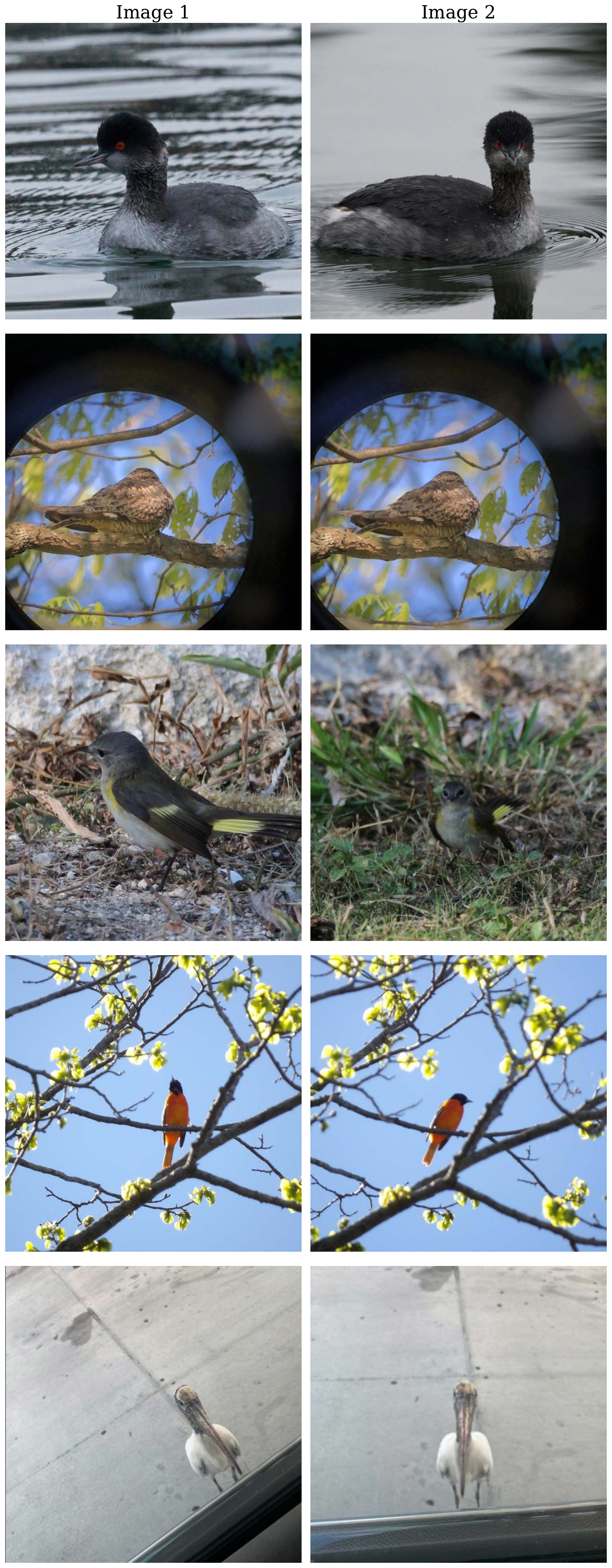}
    \caption{
        \textbf{Additional iNat-Seen examples.}
        Additional test examples from iNat-Seen, sampled randomly. 
        Images center-cropped to square.
    }
    \label{fig:supp_inat_seen}
\end{figure}

\begin{figure}[h!]
    \centering
    iNat-Unseen Additional Examples\par\medskip
    \includegraphics[width=0.45\textwidth]{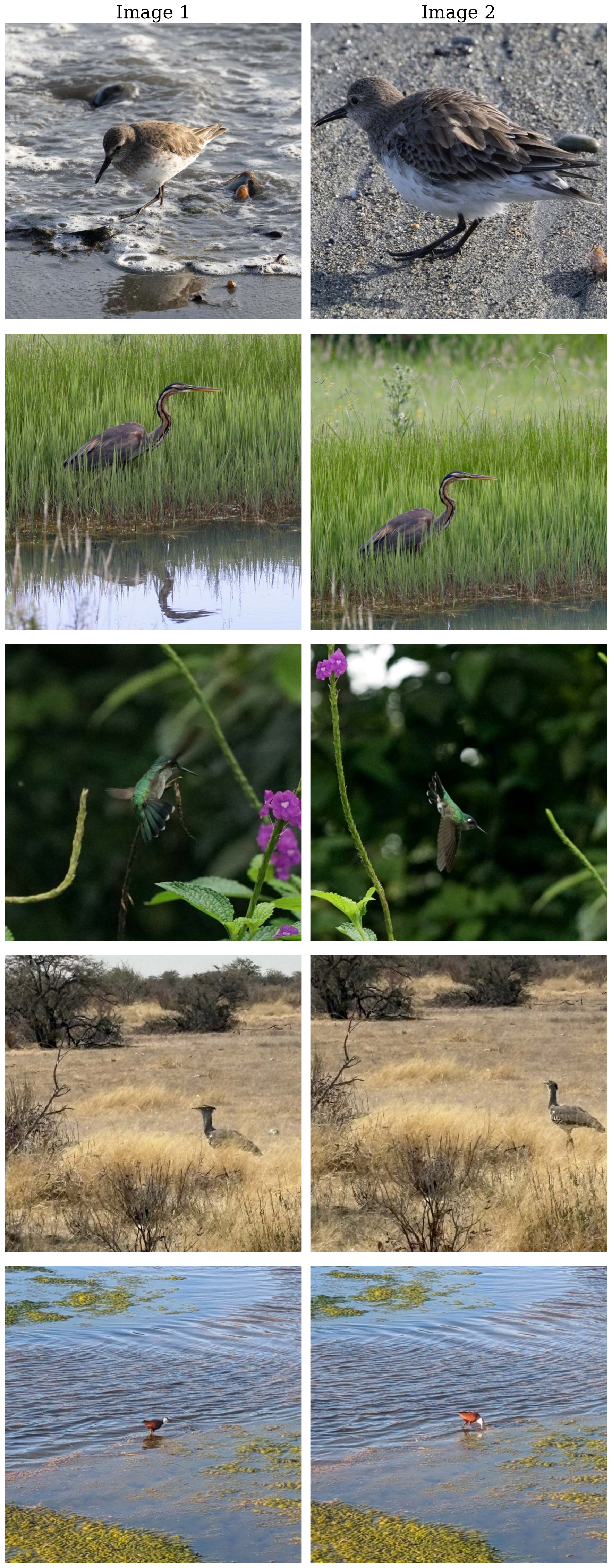}
    \caption{
        \textbf{Additional iNat-Unseen examples.}
        Additional test examples from iNat-Unseen, sampled randomly. 
        Images center-cropped to square.
    }
    \label{fig:supp_inat_unseen}
\end{figure}

\begin{figure*}[h!]
    \centering
    \includegraphics[width=0.8\textwidth]{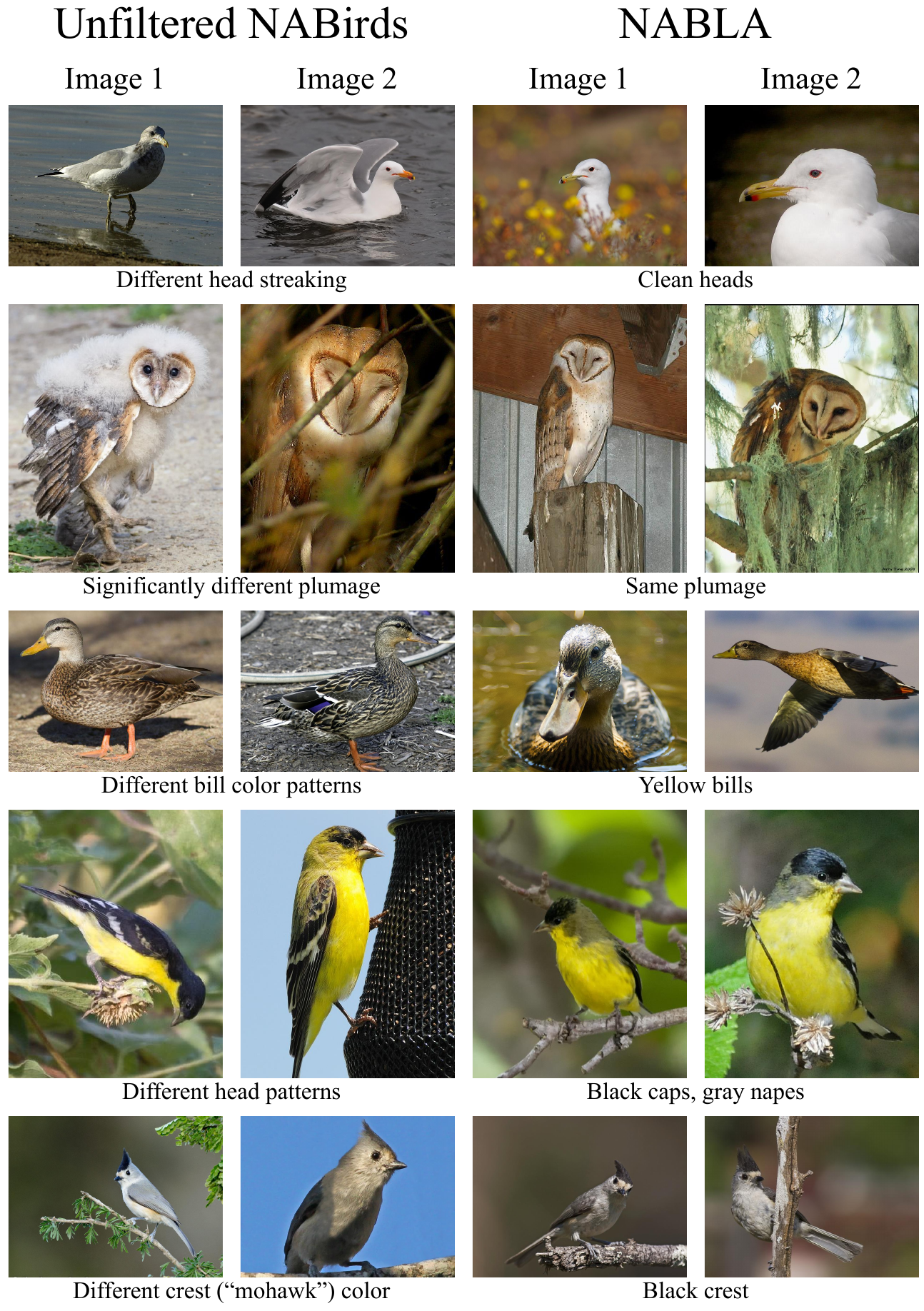}
    \caption{
        \textbf{Unfiltered NABirds vs NABLA.}
        Comparing unfiltered sample pairs from the NABirds test set to NABLA pairs of the same species.
        Unlike NABLA, pairs in NABirds can have significantly different plumages between the two images despite sharing the same class.
        This can lead to inaccurate evaluations for the identity-preserving generation task.
    }
    \label{fig:supp_mismatch}
\end{figure*}

We show additional image pairs from \datasetname, iNat-Seen, and iNat-Unseen in Figures~\ref{fig:supp_nabla}, \ref{fig:supp_inat_seen}, and \ref{fig:supp_inat_unseen}, respectively. 
In Figure~\ref{fig:supp_mismatch}, we highlight potential mismatched pairs and compare them to corresponding pairs in NABLA. 
Individual differences in plumage can occur even within a class which leads to inaccurate evaluation on the identity-preserving task.

\subsection{Other Dataset Comparisons}
\label{sec:supp_additional_datasets}
We also briefly discuss differences between \datasetname and other existing datasets which were not highlighted in the main text.
DreamBench++~\cite{peng2024dreambench} is a human-aligned benchmark but focuses on the quality of creative synthetic generations to be assessed using GPT-4o as opposed to realism and accuracy.
DeepFashion2~\cite{ge2019deepfashion2}, PODS~\cite{sundarampersonalized}, and CUTE~\cite{kotar2023these} are all true same-subject datasets but focus on everyday objects and clothes which have relatively simple pose variation.
WildlifeReID~\cite{adam2025wildlifereid} and PetFace~\cite{shinoda2024petface} are both re-identification datasets on animals. 
WildlifeReID compiles various re-identification datasets into a single benchmark but this only spans 33 species across the entire animal kingdom.
PetFace impressively spans over 250 thousand individuals across 13 families, but these photos are facially-focused for the purposes of re-identification and focuses on commonly-owned pets such as cats and dogs.
Compared to iNaturalist, we found that these two re-identification datasets were limited in terms of image resolution, pose and lighting variation, and species diversity, especially for evaluating image generation tasks.

\section{Training Hyperparameters}
\label{sec:supp_hyperparameters}
\subsection{Basic Hyperparameters}
For both OminiControl \cite{tan2025ominicontrol} and Insert Anything \cite{song2025insert} we used the Prodigy optimizer \cite{mishchenko2024prodigy} with a learning rate of 1, weight decay of 0.01, and safeguard warmup and bias correction set to true. 
On OminiControl experiments we ran with a batch size of 24 while for Insert Anything we used a batch size of 20 with gradient checkpointing for both experiments. 
We match the LoRA settings of OminiControl and Insert Anything training procedures exactly. 
At evaluation time we use guidance scale of 2.5 for OminiControl. 
\subsection{Keypoint Control}
\label{sec:supp_keypoint}
We define a sparse skeleton on the NABirds keypoints. 
The skeleton is defined with the following edges: (bill, crown), (crown, nape), (left eye, bill), (right eye, bill), (belly, breast), (breast, bill), (back, nape), (tail, back), (left wing, back), and (right wing, back). 
We explored a variety of joint and edge sizes. 
We found the best results were for a joint diameter of 15 pixels and edge width of 10 pixels.
\subsection{Proprietary Model Settings}
\label{sec:supp_prop_models}
Nano Banana and GPT-4V were run on their public websites from Nov. 1-7'25 using the ``Create Image" feature on Pro and default modes, respectively. The prompt and images are described in Figure 2 in the paper.
The propriety models were provided the subject image, masked background, and the prompt ``Please inpaint this bird into the pose given by the black mask.''

\section{Additional Results}
\subsection{Correlation Graphs}
\label{sec:supp_correlation_graphs}
Performance correlation graphs have been also been generated for the NABLA and iNat-Unseen pair and iNat-Seen and iNat-Unseen pair in Figures~\ref{fig:supp_graph_nabla_unseen} and \ref{fig:supp_graph_inat_both}, respectively.
We see the correlation between the three datasets is strong, but the iNat-Seen and iNat-Unseen datasets have the most similar performance.

\subsection{Generation Results}
\begin{figure*}[h!]
    \centering
    \datasetname Additional Generations\par\medskip
    \includegraphics[width=0.95\textwidth]{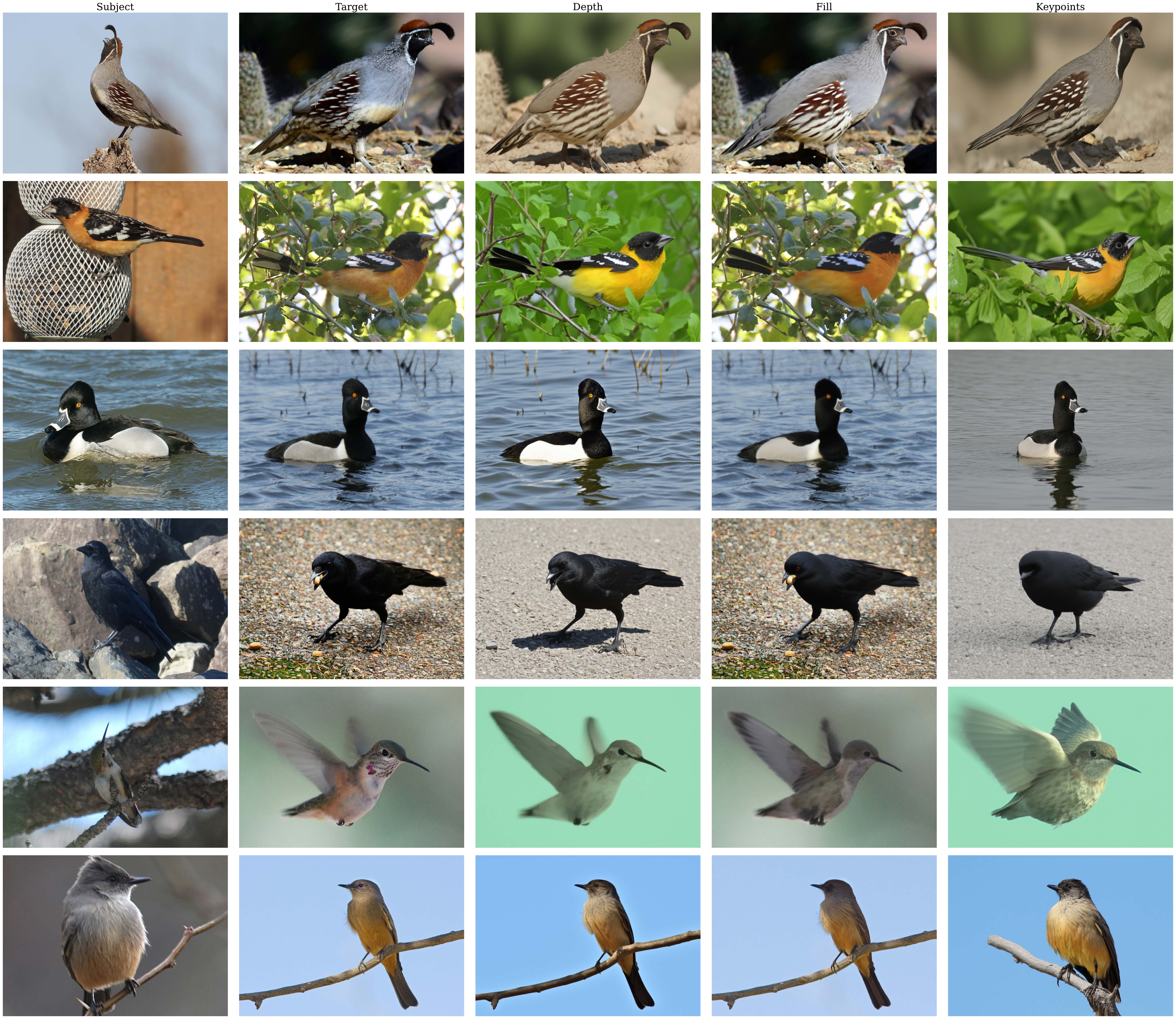}
    \caption{
        \textbf{Additional \datasetname generations.}
        Additional test examples generations from \datasetname, sampled randomly. 
        Generations from best model of each control type.
        Images center-cropped to square.
    }
    \label{fig:supp_nabla_gen}
\end{figure*}

\begin{figure*}[h!]
    \centering
    iNat-Seen Additional Generations\par\medskip
    \includegraphics[width=0.95\textwidth]{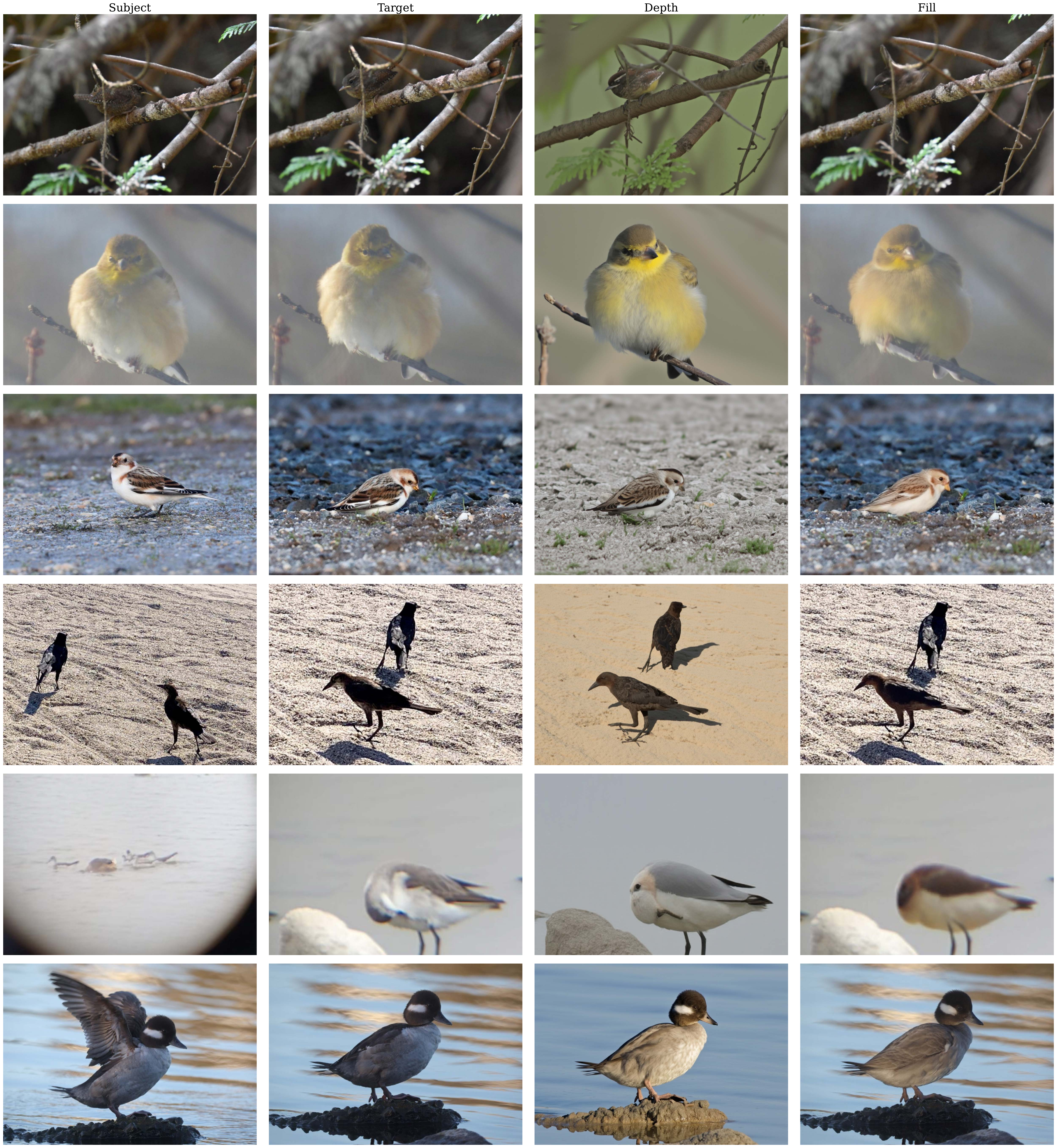}
    \caption{
        \textbf{Additional iNat-Seen generations.}
        Additional test examples generations from iNat-Seen, sampled randomly. 
        Generations from best model of each control type.
        Images center-cropped to square.
    }
    \label{fig:supp_inat_seen_gen}
\end{figure*}

\begin{figure*}[h!]
    \centering
    iNat-Unseen Additional Generations\par\medskip
    \includegraphics[width=0.95\textwidth]{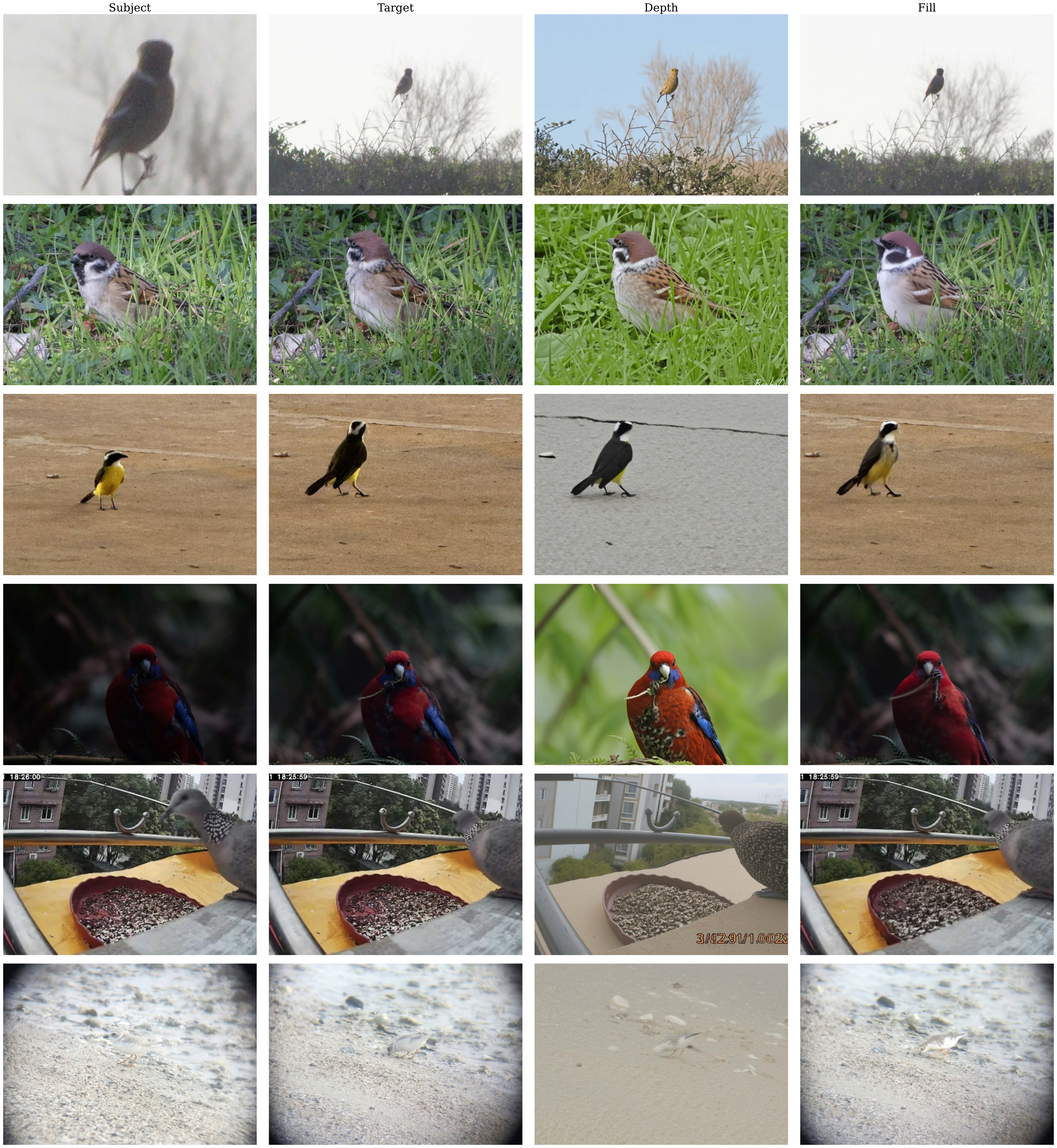}
    \caption{
        \textbf{Additional iNat-Unseen generations.}
        Additional test examples generations from iNat-Unseen, sampled randomly. 
        Generations from best model of each control type.
        Images center-cropped to square.
    }
    \label{fig:supp_inat_unseen_gen}
\end{figure*}

We show additional generation results on our best models for each control from \datasetname, iNat-Seen, and iNat-Unseen in Figures~\ref{fig:supp_nabla_gen}, \ref{fig:supp_inat_seen_gen}, and \ref{fig:supp_inat_unseen_gen}, respectively. 

\subsection{Individual-Birds Dataset Evaluation}
\label{sec:supp_indbirds}
In Table~\ref{tab:supp_indbirds}, we present results from evaluating our fine-tuned models on Individual-Birds~\cite{ferreira2020deep} from val+test sets in WildlifeReID-10k~\cite{adam2025wildlifereid}.
For each individual, 3 random image pairs were selected for evaluation. 
Since many Zebra Finch images include multiple subjects, we excluded these from our evaluation. 
We observe very similar trends to iNaturalist --- models trained on NABirds using proxy pairs show improved performance over baselines. 

\subsection{Mixed Fill + Depth Control}
\label{sec:supp_mixed}
In Table~\ref{tab:supp_mixed}, we present results after fine-tuning OminiControl + Flux-Kontext using a mixed depth + fill control mode. 
The control image is simply the fill image with the pixels from the depth map pasted into the bird mask, as in Figure~\ref{fig:supp_mixed}.
We observe very similar results between the fill, depth, and mixed models, indicating a single image combination of the two controls is insufficient for improvement. 

\begin{table}
    \begin{center}
    \begin{scriptsize}
    \begin{tabular}{r | c | c | c | c | c  }
    \toprule
    Control & Arch & DINO $\uparrow$ & SigLIP $\uparrow$ & LPIPS $\downarrow$ & MSE $\downarrow$\\
    \midrule
    Fill & Ins-A* & 0.77 & 0.93 & 0.021 & 21.6 \\
    Fill & Ins-A & 0.81 & 0.95 & 0.018 & 20.0 \\
    Fill & Om-S* & 0.28 & 0.79 & 0.050 & 22.3\\
    Fill & Om-S & 0.51 & 0.90 & 0.031 & 24.3\\
    Depth & Om-S* & 0.42 & 0.84 & 0.042 & 25.1\\
    Depth & Om-S & 0.47 & 0.87 & 0.033 & 26.6\\
    \bottomrule
    \end{tabular}
    \vspace{-3mm}
    \caption{
    Evaluating our models on random pairs in IndividualBirds from val+test sets in WildlifeReID-10k. 
    * indicates baseline.
    3 random image pairs were selected per individual. 
    Zebra Finches were excluded since many images captured multiple subjects.
    }
    \label{tab:supp_indbirds}
    \end{scriptsize}
    \end{center}
    \vspace{-6mm}
\end{table}

\begin{table}
    \small 
    \begin{center}
    \begin{tabular}{r | c | c | c | c  }
    \toprule
    Control & DINO $\uparrow$ & SigLIP $\uparrow$ & LPIPS $\downarrow$ & MSE $\downarrow$\\
    \midrule
    Depth & 0.77 & 0.93 & 0.063 & 57.0 \\
    Fill & 0.78 & 0.94 & 0.060 & 51.0 \\
    Depth + Fill & 0.77 & 0.94 & 0.061 & 57.0\\
    \bottomrule
    \end{tabular}
    \vspace{-3mm}
    \caption{Comparing the mixed control mode of depth + fill to existing results on Om-K. The results are similar to depth/fill only.}
    \label{tab:supp_mixed}
    \end{center}
    \vspace{-10mm}
\end{table}


\end{document}